# Prescriptive tool for Zero-Emissions Building (ZEB) fenestration design using hybrid metaheuristic algorithms

Rosana Caro[a], Lorena Cruz[b], Pablo S Naharro[c], Santiago Muelas[c], Arturo Martínez[a], Kevin King Sancho[c], Elena Cuerda[d], María del Mar Barbero-Barrera[a], Antonio LaTorre[e]

[a] *Department of Construction and Technologies in Architecture. Escuela Técnica Superior de Arquitectura. Universidad Politécnica de Madrid. Avenida Juan de Herrera, 4, 28040 Madrid, Spain*

[b]*Lurtis Rules SL, Plaza del Sol, 31, 28938 Móstoles, Madrid, Spain*

[c] *Lurtis Rules LTD, Wood Centre for Innovation, Quarry Road, Oxford, United Kingdom*

[d]*Department of Architecture. Escuela de Arquitectura. Universidad de Alcalá de Henares. Calle Santa Úrsula, 8. Alcalá de Henares. 28801 Madrid, Spain.*

[e] *Center for Computational Simulation (CCS), Universidad Politécnica de Madrid. Campus de Montegancedo. Avenida de Montepríncipe s/n. 28660 Boadilla del Monte, Madrid, Spain.*

## Abstract

Designing Zero-Emissions Buildings (ZEBs) involves balancing numerous complex objectives that traditional methods struggle to address. Fenestration, encompassing façade openings and shading systems, plays a critical role in ZEB performance due to its high thermal transmittance and solar radiation admission. This paper presents a novel simulation-based optimization method for fenestration designed for practical application. It uses a hybrid metaheuristic algorithm and relies on rules and an updatable catalog, to fully automate the design process, create a highly diverse search space, minimize biases, and generate detailed solutions ready for architectural prescription. Nineteen fenestration variables, over which architects have design flexibility, were optimized to reduce heating, cooling demand, and thermal discomfort in residential buildings. The method was tested across three Spanish climate zones. Findings provide valuable guidance for ZEB design, highlighting challenges in reducing cooling demand in warm climates, and showcasing the superior efficiency of automated movable shading systems compared to fixed solutions.

*Key words*: Zero emission building; Efficient building envelope; Fenestration optimization; Multi-objective optimization; Hybrid metaheuristic algorithms; Computer-aided building design; Building shading system

# 1. Introduction[1]

Since 2020, all new EU buildings must meet the "Nearly Zero-Energy Buildings" (NZEB) standard until 2030 to support carbon neutrality by 2050 [1]. From 2030 onward, stricter rules apply, requiring a 10% lower primary energy demand under the "Zero-Emission Buildings" (ZEB) standard set in the 2024 revision of the Energy Performance of Buildings Directive (EPBD) [2], [3]. The 2010/31/EU Directive [4] defined a NZEB as a building with a very high energy performance where the nearly zero or very low amount of energy still required should be covered to a very significant extent by energy from renewable sources, including those produced on-site or nearby. Since then, there has been a proliferation of new definitions, [5], [6] and the research field on the topic is now considered to be highly consolidated [7], [8].

However, achieving zero emissions is challenging, requiring a shift from traditional design methods, as ZEB standards have evolved beyond operational efficiency to balance life-cycle impact, Indoor Environmental Quality (IEQ), and economy. The current process depends on specialist collaboration and on the results of several evaluative, non-prescriptive simulation tools, requiring continuous trial-and-error to reach a compromise. This traditional approach is inefficient, costly, time-consuming, with often inaccurate performance predictions [9], [10]. It is now acknowledged that building simulation programs, when used independently by different specialists (operational energy performance, IEQ, life cycle analysis (LCA),

---

[1] **Abbreviations**: CI: Compactness index ($m^3/m^2$); COP: Coefficient Of Performance; EC: Energy consumption (operational) ($kWh/m^2 \cdot year$); ECh: Energy consumption for heating ($kWh/m^2 \cdot year$); ECc: Energy consumption for cooling ($kWh/m^2 \cdot year$); ECl: Energy consumption for lighting ($kWh/m^2 \cdot year$); ED: Energy demand ($kWh/m^2 \cdot year$); EDh: Energy demand for heating ($kWh/m^2 \cdot year$); EDc: Energy demand for cooling ($kWh/m^2 \cdot year$); EUI: Energy Use Intensity ($kWh/m^2 \cdot year$); GA: Genetic algorithm; MOO: Multi-objective optimization; LCC: Life Cycle Cost (€/$m^2$); LCEI: Life Cycle Environmental Impact; PEC: Primary energy consumption ($kWh/m^2 \cdot year$), n50: Air change rate at 50 Pa ($h^{-1}$); NCT: Not comfort time (h); NEC: National Energy Code, NV: natural ventilation; OF: Objective function; PCM: Phase change material, PV: Photovoltaic power; SC: Solar control, SHGC: Solar Heat Gain Coefficient (dimensionless), VT: Visible transmittance (dimensionless); U: Thermal transmittance ($W/m^2 \cdot K$); WWR: Window to wall ratio (dimensionless); σ-value: Standard deviation; λ: thermal conductivity ($W/m \cdot K$); δ: density ($kg/m^3$)

life cycle cost (LCC)) do not suffice to meet the ambitious targets pursued by many policies and "need to undergo an innovation jump towards a more holistic tools and workflows that address the new requirements brought forward by the increased complexity [11], [12], [13] ". In fact, ZEB design can be addressed as an optimization problem. Optimization means finding the best solution(s) among different feasible alternatives, where feasible solutions mean those that satisfy all the constraints [14].

Radford and Gero laid the theoretical groundwork for applying computational techniques to architectural optimization in 1980 [15]. Since then, Artificial Intelligence (AI) has become prevalent in architecture, within optimization methods using Genetic Algorithms (GA) emerging as the most popular method [16]. Optimization methods are categorized as exact or approximate [17]. Exact methods guarantee optimal solutions but can be slow and unsuitable for problems where the gradient cannot be calculated. Consequently, interest has grown in approximate methods, which seek "good" solutions within a reasonable timeframe, even if they do not guarantee optimality. Good solutions are considered those that achieve favourable results for the different objectives and, at the same time, meet the regulatory limit for the selected indicators. Approximate methods include heuristics like Local Search, which iteratively improves solutions within a neighborhood, and metaheuristics [18], introduced by Glover [19], which combine different heuristic methods in a higher-level framework with the purpose of efficiently and effectively exploring the search space. Among metaheuristic methods, Evolutionary Algorithms (EAs), inspired by natural evolution, are particularly notable. Holland's 1975 approach [20] pioneered their use in complex architectural optimization, while Storn and Price's 1995 Differential Evolution (DE) algorithm [21] proved especially successful. Initially designed for the Chebyshev polynomial fitting problem, DE optimizes a problem by managing a population of candidate solutions and generating new candidates by combining existing ones according to its simple formulae. Since its inception in 1995, DE has drawn the attention of many researchers all over the world resulting in a lot of variants of the basic algorithm with improved performance. One such variant is the SHADE-ILS [22] algorithm, which combines the Success-History based Adaptive Differential Evolution (SHADE) algorithm [23], a DE variant that dynamically adapts its parameters, with two local searches: the L-BFGS-B and MTS local search [22]. This hybrid algorithm outperformed others in the CEC benchmark for complex high-dimensional problems [24], particularly excelling in the most challenging scenarios. As a result, SHADE-ILS proves to be a promising candidate for addressing the complexity inherent in architectural design.

In optimization building applications, the highly diverse energy-related parameters are often addressed separately. Barber and Krarti [25] revision focused on research optimizing the energy systems of buildings (HVAC), while that of Kheiri [26] focused on the building envelope. The building **envelope** separates interior spaces from the external environment, regulating heat, light, moisture, and noise exchange. It is composed of both opaque and transparent surfaces arranged in a specific configuration. Beyond air transfer from infiltration and ventilation, operational energy performance depends heavily on envelope properties [27]. An efficient envelope design minimizes energy flow, reducing energy demand (ED), system capacity, costs, and environmental impact.

**Fenestration**, the most thermally inefficient part of the envelope [28], includes windows and skylights that provide ventilation, daylight, and views but also contribute to overheating and glare due to high thermal transmittance and solar gain admission. Fenestration consists of three components: (1) glazing (made of glass or plastic), (2) framing (mullions, dividers, mounting bars), and (3) shading devices (fixed or movable) [29]. Glazing and frames are static, while weather and solar radiation vary, making it difficult to balance heat, daylight, and views while preventing glare and overheating. Movable shading devices, like blinds or curtains, help maintain this balance [30]. The initial investment for ZEB fenestration is significantly higher than for opaque envelope surfaces. According to 2018 data from the EU CRAVEzero project [31], the average cost of insulated walls in Spain is 81 €/m², while double-glazed windows cost 307 €/m². Despite its critical role, fenestration decisions are often made early in design using qualitative methods shaped by professional inertia, leading to potential cost overruns or malfunctions. Altering fenestration post-construction is costly, and climate change adds new challenges. While multi-objective optimization could greatly improve fenestration design, no practical tools currently exist to automate this process for ZEBs.

## 1.1. Objective

As things stand, this article introduces a novel simulation-based optimization method that uses a hybrid metaheuristic algorithm and conducts energy simulations by calling the Energy Plus simulator [32] to automate fenestration design in residential ZEBs. The method streamlines the design process, providing precise fenestration solutions tailored to project specifications, room variations, and façade solar orientation, making it a powerful prescriptive tool. To validate its effectiveness, the method was applied to a four-floor apartment under reference weather conditions in three Spanish climate zones.

Nineteen fenestration variables critical to ZEB energy performance were optimized to minimize heating and cooling demands and thermal discomfort, while maximizing material savings. The method optimizes the window width and height for each room, and for each façade, it optimizes the double glazing composition, frame material, automated shading programs, and presence and dimensions of overhangs and fins. Specialized solutions like dynamic smart windows and photovoltaic or phase change material integrated glazing were excluded for practical applicability. Although standard operating conditions were assumed, the method can be adapted to other operation scenarios.

## 1.2. Literature review

The analysis of 23 articles from the Scopus Database (1997-2024) identified two primary research approaches. The first, termed Predefined Solutions (PDS), begins with manually pre-designed configurations treating physical properties like thermal transmittance (U-value), solar transmittance (SHGC), and air change rate at 50 Pa (n50) as variables that can only vary within a limited set of discrete values. The second, Rule-based Generation of Components (RGC), uses information banks and rules to automatically generate multi-layered building components. Optimization methods were categorized into: Parametric Design (PD) techniques, which represent a design through parameters (e.g., BIM) without relying on algorithms [33], and algorithm-based methods. A chronological list of the articles, including research approach, optimization method, case study type, optimization objectives and variables, simulation software, and optimization algorithm, is provided in **Table 1.**

**Table 1**

**Literature review**

| Ref. | Approach | Method | Case study | Objectives | VARIABLES ||||
|---|---|---|---|---|---|---|---|---|
| | | | | | Orientation and building shape | Walls and/or roof | Fenestration | Active systems and/or operation |
| 1997 Marks [33] | PDS | Parametric | Not specified | Material cost; ECh | Compactness[1]; Aspect ratio[2] | | | |
| 2002 Caldas&Norfold [34] | RGC | Micro GA [35] | Offices | ECh; ECc; ECl | | | Window width and/or height | |
| 2002 Coley and Schukat [36] | PDS | GA | Residential | ECh; Ecc | Solar orientation; Volume | Types of walls and/or roofs | WWR; Window position | |
| 2003 Werner and Mahdavi [37] | PDS | Parametric | Residential | EDh | Solar orientation; Volume; Compactness | | WWR | |
| 2005 Wang et al. [38] | PDS | GA | Offices | LCC; LCEI | Solar orientation; Aspect ratio | Types of walls and/or roofs | WWR | |
| 2009 Wright and Mourshed [39] | RGC | GA | Commercial | EC | | | WWR; Window position; Window width and/or height | |
| 2013 Shi and Yang [40] | RGC | GA | Not specified | EC; Daylighting; Photovoltaic electricity production | | | Window width and/or height | PV roof area |
| 2014 Manzan [41] | RGC | GA (ModeFRONTIER) | Offices | PEC | | | Shading device material and/or geometry | |
| 2015 Weng et al. [42] | RGC | GA + Energy Plus | Not specified | EDh | Internal partitions; Compactness | | | |

**Table 1**
Literature review (continued 1)

| | | | | | VARIABLES | | | |
|---|---|---|---|---|---|---|---|---|
| Ref. | Approach | Method | Case study | Objectives | Orientation and building shape | Walls and/or roof | Fenestration | Active systems and/or operation |
| 2015 Méndez Echenagucia et al. [43] | PDS | NSGA-II + Energy Plus | Office | ECh; ECc; ECl | | Thickness of external walls | WWR; Window position; Window width and/or height; Glazing type | |
| 2015 Carlucci et al. [44] | PDS | GA (GenOpt) + Energy Plus | Residential | Thermal and visual comfort | | Types of walls and/or roofs | Type of glazing | Natural ventilation; Shading operation |
| 2016 Konis et al. [45] | RGC | GA Grasshopper-Octopus | Commercial | Daylighting, solar control, natural ventilation | Volume | | WWR; Shading device material and/or geometry | |
| 2016 Vera et al. [46] | PDS | PSO-HJ (GenOPT)+ mkSchedule +Radiance +EnergyPlus | Offices | Visual comfort; ECh; ECc; ECl | | | Shading device material and/or geometry | |
| 2019 Fang and Cho [47] | PDS | GA (Octopus) + Ladybug-Honeybee+Radiance+Energy Plus | Office | EC; Daylighting | Aspect ratio | | Window width and/or height; Shading device material and/or geometry; Skylights form and/or composition | |
| 2019 Zhai et al. [48] | RGC | NSGA-II + EnergyPlus | Office | ED; Thermal comfort; Daylighting | | | WWR; Double-glazing composition (glass and gas) | |
| 2019 Ascione et al. [49] | PDS | NSGA-II + ANN + Energy Plus | Office | ED; Thermal comfort | Solar orientation; Aspect ratio$^2$ | Walls thickness; Walls' renders colors; λ; δ | WWR; Type of glazing; Shading device material and/or geometry | Heating and Cooling set-point temperatures |
| 2020 Ciardello et al. [50] | PDS | aNSGA-II | Residential | EC; Costs; $CO_2$ | Solar orientation; Compactness; Aspect ratio; | | WWR | |
| 2020 Zhao and Du [51] | PDS | NSGA-II + Design Builder | Offices, hotel | ECh; ECc; Ecl; Thermal comfort | Solar orientation | | Type of glazing; Type of glass; Shading device material and/or geometry | |

**Table 1**
Literature review (continued 2)

| | | | | | VARIABLES | | | |
|---|---|---|---|---|---|---|---|---|
| Ref. | Approach | Method | Case study | Objectives | Orientation and building shape | Walls and/or roof | Fenestration | Active systems and/or operation |
| 2023 Kamazani and Dixit [52] | PDS | NSGA-II | Office | EC; Embodied energy; Operational and embodied carbon | Solar orientation | Wall composition | WWR; Type of glass; Type of gas | |
| 2024 Luo et al. [53] | PDS | Non-dominated sorting GA of Wallacei | Parking building | EC, Thermal comfort; Cost | | Type and thickness of insulating materials in walls and roof | WWR; Types of windows (x17) | PV on envelope (area, material and position); Types of PV material (x2) |
| 2024 Benaddi et al. [54] | PDS | GENOPT | Educational | LCC; $CO_2$; Thermal comfort | | Retrofit solutions for walls (x4) and roof (x3); Thickness and position of insulation in walls | WWR; Types of glazing (x4) | |
| 2024 Mehrpour et al. [55] | PDS | NSGA-II | Residential | EDh; EDc; EC | Solar orientation | Wall type (x7); Roof type (x2); PCM in walls (x4) | WWR; Types of glazing (x6); Exterior and interior shading | |
| 2024 Pan et al. [56] | PDS | Deep reinforcement learning | Educational | EC, Thermal comfort; $CO_2$ | | Wall U-value (x1); Floor U-value (x2); Infiltration rate (x3) | WWR (x6); Glazing U-value (x4) and SHGC (x5); Window open rate (x7) | Heating COP (x7); Cooling COP (x8); Heating, Cooling and NV set-point temperatures (x10, x11, x12) |

[1] Compactness: total volume of a building or space to its outer surface ratio in $m^3/m^2$

[2] Aspect ratio: relation between the width and the length of a rectangular floor plan

1.3. Research gaps and novelty

The literature review underscored fenestration's vital role in energy efficiency, with 21 of 23 studies incorporating at least one fenestration variable. It also showed that genetic algorithms (GA) have largely replaced parametric design techniques. However, several research gaps remain.

A tendency to oversimplify the problem was identified in the selection of sub-components, variables, and optimization objectives, which may not always be optimal. This can affect practical applicability, even when advanced algorithms are used. For instance: Window frames were often overlooked, despite their significant impact on performance and cost [57]; Only four of 23 studies optimized glazing dimensions (width and length), with most relying on the Window-to-Wall Ratio (WWR) index, which require additional manual calculations by architects to meet design needs; Although many studies considered movable shading materials and geometry, only one included automated shading control as a variable, despite its crucial role in bridging the performance gap [9], [58]; Variables of very different nature (envelope, human behavior, HVAC) were often mixed, complicating result interpretation due to complex interactions; Thermal properties like U-values and n50 were commonly used as variables, but offer limited practical value for architects and engineers, who must adhere to normative limits; The Predefined Solutions (PDS) approach was dominant (16 papers) and often relied on region-specific materials, which limited broader applicability, constrained construction options, and lacked the detailed information architects need for precise prescriptions; Energy consumption (EC) was prioritized over energy demand for heating and cooling (EDh, EDc) in envelope optimization, even though EC is more closely linked to the efficiency of a building's HVAC system; Thermal comfort was considered as optimization objective in only 7 studies.

As a result, the proposed solutions often lacked the high-resolution detail necessary for precise decision-making in architectural projects, and true automation (eliminating the need for manual calculations) was not achieved. Additionally, most studies focused on high-energy-consuming public buildings, using energy consumption and emissions as key optimization factors. However, residential buildings, which represent the majority of structures worldwide, make up only 23% of the studies. In Spain, for instance, residential buildings account for 66.1% of the total building stock [59].

In contrast, the proposed method introduces several innovations:

- It comprehensively addresses the problem's complexity, selecting optimization objectives tied directly to fenestration performance, based on key regulatory indicators and widely accepted evaluation systems. Nineteen variables, offering architects and engineers significant decision-making flexibility, are chosen, with regulated ones treated as constants.
- A fully automated workflow integrates simulation tools, streamlining the process, removing manual intervention, and enhancing applicability.
- It automatically generates all possible fenestration sub-component combinations using a custom catalog of market products and a set of rules derived from expert knowledge, energy transfer principles, and building regulations, contrasting with manual approaches that restrict solutions variability and introduce bias.
- Designed for practical use, it delivers detailed solutions for architectural specifications, accounting for room variations and façade orientation, facilitating easy integration into project documents. This level of detail surpasses other studies, where solutions typically only "suggested" optimal values for properties like U-value, SHGC, and WWR, without fully specifying all variables values of the optimized sub-component.

## 2. Methods

The tool streamlines the refinement of fenestration designs through an optimization process driven by a hybrid evolutionary algorithm. Using five primary inputs: a catalog of window components, a rule set defining design constraints for glazing and shading dimensions, regulatory values, weather data, and the model geometry, the tool generates an initial pool of potential solutions. These solutions are subsequently refined through an iterative optimization process focused on exploration and exploitation, employing the SHADE+L-BFGS-B algorithm. To assess the quality of the solutions, the tool adopts a multi-criteria evaluation framework that integrates several analyses and simulations: (a) energy performance (simulated by conducting an EnergyPlus simulation), (b) thermal comfort (simulated also via EnergyPlus), and (c) material minimization for fenestration components (a simplified cost metric). The overall workflow is illustrated in Fig. 1

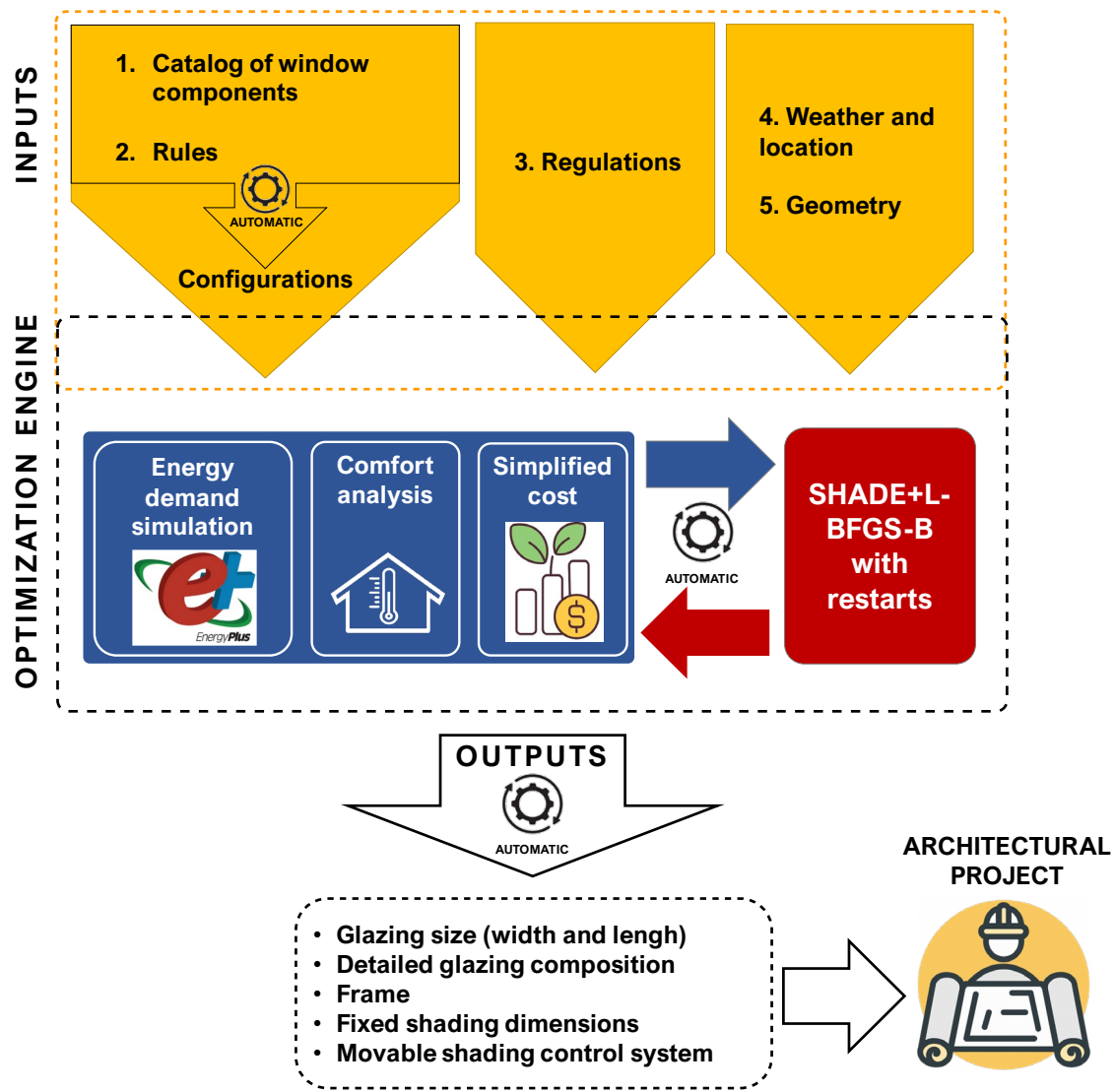

**Fig. 1**. Optimization tool workflow

The research followed four main steps, outlined below to clarify the workflow.

1. **Problem formulation** (Section 2.1). Defined the optimization objectives (what we do want to achieve) and their evaluation criteria, the fenestration components selected for optimization and the parametric model better suited for the objectives.
2. **Optimizer development.** (Section 2.2). Detailed the fitness function (satisfaction regions, the weighting and normalization of objectives, the quality and penalty functions), the two optimization algorithms employed, the encoding space, the experimental scenario, and the hyperparameter selection.

3. **Case study and energy model set**-up (Section 3). Defined the building's geometry, operational conditions, regulatory framework, and locations and climatic conditions considered in the simulations.

4. **Assessment of the optimization process** (Section 4.1). The performance of the SHADE+L-BGFS-B algorithm used in this research was compared against the Genetic Algorithm (GA), commonly employed in the literature for similar optimization problems.

5. **Assessment of the optimized solutions** (Section 4.2). Two window shading scenarios were evaluated: S1, where both fixed and movable shading for all windows was available for the optimization, and S2, where only fixed shading was available. For each scenario and location, 15 optimization runs were conducted. The top 10 solutions per run, based on EDh, EDc, and NCT results (150 solutions in total), were downloaded into a spreadsheet for analysis.

2.1. Problem formulation.

Optimization methods rely on three key inputs: objective functions (OF), variables, and parameters. OFs act as performance metrics [60], evaluating solutions against the established goals. Variables represent adjustable attributes of building elements, with optimization identifying the best values within a defined range. Parameters are fixed attributes, like location and weather, that remain constant during simulations. In this study, factors like solar orientation, urban context, building shape, volume, and internal gains were treated as parameters due to the limited control designers have over them. Each combination of variable values forms a potential solution, and all solutions together define the optimization search space.

*2.1.1. Objectives of the optimization and their evaluation*

When designing high-performance buildings, developers prioritize objectives based on their regulatory, socio-economic, and cultural contexts. The EU's energy strategy [61] [62] upholds the "energy efficiency first" principle, requiring efficiency in major policies and investments. In buildings, this means reducing energy demand, cutting fossil fuel use, and increasing local renewables. However, ZEB expectations now extend beyond efficiency. Green certifications and international guidelines incorporate broader societal values. The 2024 EPBD [3] stresses balancing energy efficiency with cost, circularity, health and comfort, indoor environmental quality and the improved adaptive capacity of the building to climate change.

Considering this holistic approach, three optimization objectives were selected for this research.

1. minimizing the annual energy demand for heating and cooling (EDh and EDc in kWh/m$^2$),
2. minimizing the number of Not Comfort Time per year (NCT in hours)
3. minimizing the amount of material used for the construction of the fenestration components (simplified cost)

To assess the first two objectives, the proposed approach conducts an Energy Plus simulation for each solution generated throughout the optimization process and obtains the corresponding value from the simulation output.

The first objective aligns with the EU's energy efficiency targets, as heating and cooling account for half of the EU's final energy consumption [63]. Noticeably, the use of energy for space cooling is growing faster than for any other end use in buildings, more than tripling between 1990 and 2016 [64]. Although thermal comfort is not mandatory, it influences energy consumption and health, being linked to Sick Building Syndrome [65]. The third objective focuses on reducing economic and environmental costs by minimizing materials in the fenestration system while maintaining performance. The evaluation criteria for objectives 1 and 2 is outlined below:

**Energy demand evaluation.** The current Spanish National Energy Code (NEC) [66] restricts annual primary energy consumption for new buildings but no longer imposes limits on EDh or EDc since 2019. This shift prioritizes HVAC efficiency over passive systems like building envelopes and shading. However, to achieve a viable ZEB, energy supply and demand must be balanced, minimizing energy demand with a high-performance envelope to reduce HVAC capacity needs and enable optimized on-site renewable energy generation [6]. Therefore, in the absence of regulatory limits, energy demand in Spain was evaluated using three references: 2009 statistical values for existing housing [67]; the NEC prior to 2019 [68], and (3) BREEAM [69] and Passive House [70] certification criteria. Then, EDc upper limits were set at 10, 15, and 20 kWh/m²·year for León, Madrid, and Sevilla, respectively, while those for EDh were 46, 30, and 12 kWh/m²·year for the same cities.

**For thermal comfort,** no universal evaluation criteria exist. This study followed EN 16798-1:2019 [71], which replaced EN 15251:2008 [72]. EN 15251 recommended that deviations in indoor environment parameters should not exceed 3-5% of the study period. Accordingly, the "acceptance" level for Not Comfort Time (NCT) was set at 3-5%, "good" at 1-3%, and "high quality" at 0-1% of total annual occupied hours. Since the NEC does not account for periods without occupants, the upper acceptance limit for NCT was set at 438 hours per year (5% of 8760).

The NCT index was calculated using the simple method of ASHRAE Standard 55 [73] in Energy Plus [74]. It defines comfort ranges for 80% occupant acceptability, applicable in spaces with air speeds below 0.20 m/s, metabolic rates between 1.0 and 1.3 met, and clothing insulation between 0.5 and 1.0 clo. Despite its limitations [75], this model was chosen due to its alignment with EU energy regulations prioritizing climate-controlled spaces over adaptive comfort approaches.

*2.1.2. Components of fenestration included in the optimization*

Fenestration consists of three components: Glazing, Frame and Shading devices, each with sub-components that can be optimized for specific functions [28]. The glazing and frame group is commonly referred to as the "window" (**Fig.** 2).

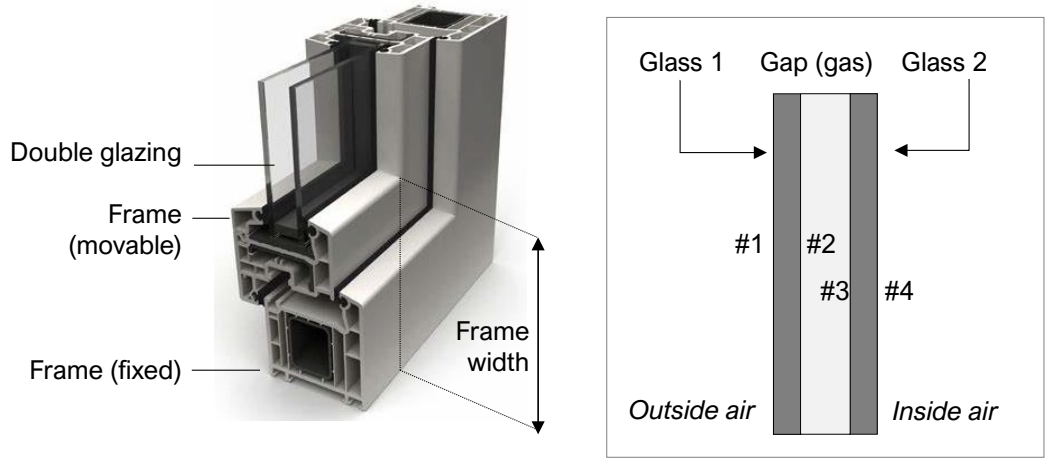

**Fig. 2**. On the left, 3D view of a window section showing its two main components: the double glazing and the frame (with a movable and a fixed part). On the right, vertical section of the double glazing with the identification of their parts and the designation of their four faces. The faces are identified by a hashtag followed by a number, ranging from the outermost face (#1) to the innermost face (#4).

The optimization focused on the following features of windows: glazing dimensions (width and height), glazing composition, and frame material. For broad applicability, the study excluded triple and single glazing, used only rectangular windows, standardized frame width (7 cm) and omitted the presence of dividers. For shading devices, the study concentrated on two highly effective systems:

- Movable shading, consisting of exterior aluminum louvered blinds with flat, equally spaced horizontal slats parallel to the glazing (**Fig. 3**). The slats can rotate between 0º (parallel and outward-facing), 90° (perpendicular), and 180° (parallel and inward-facing) to control solar radiation. The blind can be fully retracted or activated, covering the entire glazed area, with partial coverage not considered. The control system (SC) manages slat retraction, activation, and angle adjustments at each time step. The optimization variables were the slats' solar reflectance and the SC.

- Fixed shading, comprising overhangs and fins designed based on predefined façade-specific rules (Appendix C). These elements were assumed perpendicular to the façade, frame-aligned, and sharing the frame's reflectance with no heat capacity. The optimization variables included their presence, depth and extension (see **Fig.** 4).

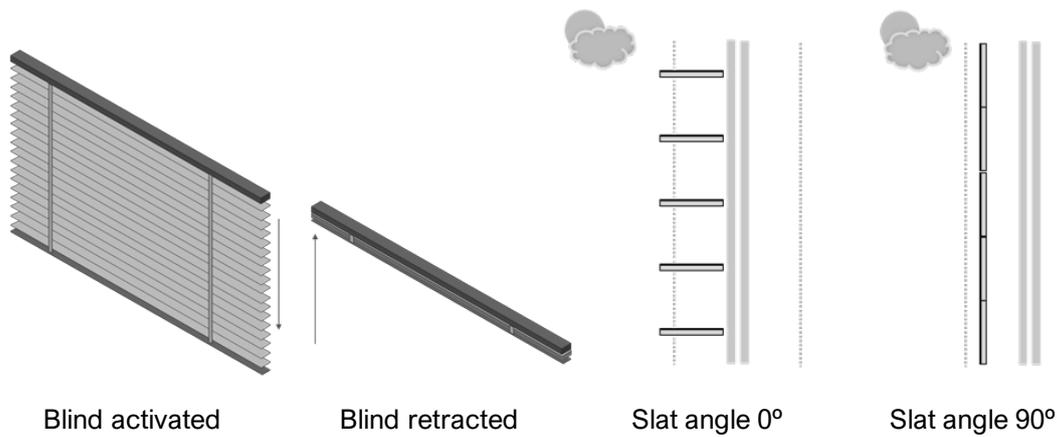

**Fig. 3.** On the left, the exterior louvered blind is shown in its two allowed positions: activated or retracted. On the right, the slat rotation angle convention is illustrated. When activated, slats can rotate between 0° (horizontal) and 90° (vertical) to block solar radiation.

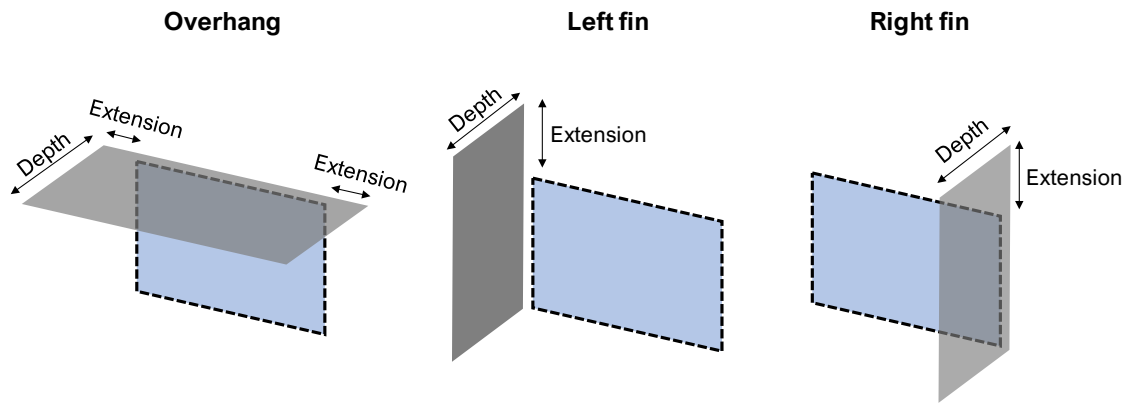

**Fig. 4.** Fixed shading surfaces considered for optimization are shown in gray: overhangs (horizontal) and side fins (vertical). Window glazing is depicted in blue. The optimized dimensions include depth and extension.

*2.1.3. Parametric model*

A parametric building model is an abstract representation of it in which some of its elements have attributes that are fixed or static, no time-dependent during simulation and therefore not subject to optimization (parameters), and some others have attributes that can vary (variables) [76]. Classifying attributes into parameters and variables, and setting their fixed values and ranges, helps form the "parameter space" by narrowing down possible states to those most relevant to the problem at hand. In this research the following attributes were considered parameters:

- **Locations, weather and geometry.** Three Spanish cities, León, Madrid, and Sevilla, representing the country's climate variability were selected. Hourly time series corresponding to the Typical Year weather data (TY) [77] for each city were downloaded from METEONORM [78] and used as input for the simulations. The building geometry represents a fourth-floor, two-bedroom apartment in a multi-family block (Section 3).
- **Fabric.** The walls had a thermal transmittance (U-value) of 0.22 W/m²·K, aligning with the strictest NEC [66] value for NZEB performance. Floors, ceilings and walls between different dwellings were adiabatic, assuming similar internal gains from adjacent units. Internal partitions, made of gypsum plasterboard and mineral wool, had a heat capacity of 11.729 kJ/m²·K and were modeled as an exposed surface area of 109.65 m². n50 was set at 0.6 ach$^{-1}$, matching the Passive House standard [70].
- Internal gains were calculated assuming four occupants (117 W/person [29]) and peak loads of 4.4 W/m² for lighting and electrical equipment [66]. Radiant fractions were 0.30 (occupancy), 0.42 (lighting), and 0.20 (equipment), with hourly profiles from NEC [66].
- The HVAC system was modeled as an Ideal Loads Air System, maintaining 22°C for heating and 25°C for cooling. It integrates heating, cooling, and ventilation via a direct expansion unit, supplying outdoor air independently of thermal power. Air is treated to meet indoor needs without added demand, with a constant flow rate of 0.024 m³/s per NEC [79]. Natural ventilation operates as an automatic free-cooling system, increasing airflow up to 4 ach$^{-1}$ [80] when outdoor temperatures are lower, preventing overheating year-round without modeling complex manual window operations. Forced ventilation "on demand" was not considered.

Table I in Supplementary Material presents the list of parameters and their selected values for optimization.

Regarding the **variables**, their precise identification, as well as their range values, is sometimes done through sensitivity analysis or other mathematical techniques, as a previous step of the application of AI algorithms [81]. In this research, literature analysis, regulations and expert knowledge were used to inform the selection process [82], [83], through interdisciplinary work involving computer scientists, research building engineers and architects. The 19 selected variables and their respective value range are listed in the following **Table 2**.

**Table 2**
Variables for optimization

|  | **Value range** |
|---|---|
| Glazing height and width (m) | According to rules (Appendix B) |
| Frame material | Wood (high or low density), Wood-aluminium, Aluminium, Vynil |
| Frame solar reflectance | 0.29 - 0.85 |
| Gas type | Air, Argon |
| Gap width (mm) | 6, 8, 10, 12, 16 |
| Thickness of glass (mm) | 4, 6, 8, 10 |
| Glass. Solar transmittance at normal incidence ($T_{sol}$) | 0.23 - 0.82 |
| Glass. Infrared Hemispherical Emissivity (emis) | 0.038 - 0.857 |
| Glass. Visible Transmittance at normal incidence ($T_{vis}$) | 0.35 - 0.90 |
| Blind. Slat solar reflectance | 0.29 - 0.85 |
| Blind. Operation programme | SC0, SC1, SC2, SC3, SC4, SC5, SC6 |
| Overhang and side fin presence | Yes/No |
| Overhang and side fin depth (m) | 0.20-1.50 |
| Overhang and side fin extensions (m) | 0-0.30 |

*2.1.4. Search space*

The Rule-based Generation of Components (RGC) approach was employed in this research. A first set of rules (see Appendix A), was created to automatically generate all possible glazing compositions from a specially designed window products catalog. These rules, based on energy transfer principles and practical experience of building design in Mediterranean climates, accounted for the various solar orientations and different rooms. Their application to the catalog produced 1,350 double-glazing solutions; including frames expanded this number to 13,500 solutions. For each combination, pyWinCalc [84] calculated the window's thermal transmittance ($U_w$), visible transmittance (VT), and Solar Heat Gain Coefficient (SHGC). A second set of rules (Appendix B) guided glazing sizing based on room type, daylight access, and furnishing practicality in multi-family dwellings. A third set (Appendix C) defined overhang and side fin dimensions (see **Fig. 4**).

The catalog derived from the International Glazing Database (IGDB) [85], was refined to remove redundant entries and focus on relevant materials. It included 44 glass types (4, 6, 8 or 10 mm thick), 13 frame options (various materials and conductance values), and 10 gap variations (air or argon, 6, 8, 10, 12 or 16 mm thick) (see **Table 2**). Non-spectrally selective laminated, tinted, or highly reflective glasses that cause a "mirror effect" or reduce indoor light quality were excluded. Selected glasses had solar and visible transmittance of 0.23–0.82 ($T_{sol}$) and 0.35–0.90 ($T_{vis}$). Low-emissivity (emis) glass was included for its energy-saving benefits in the heating season. The WPC was compiled using proportional groups of glasses from five categories:

1. Clear, uncoated glass
2. Low $T_{sol}$ glasses (0.23–0.54) without low-emissivity coating (emis = 0.84)
3. Spectrally selective glass with high $T_{vis}$ (0.56–0.68), low $T_{sol}$ (0.25–0.54), and no low-emissivity treatment
4. High $T_{sol}$ glass (0.60–0.82) with low-emissivity treatment on some of their sides (emis = 0.07–0.30)

5. Low $T_{sol}$ glasses (0.26–0.55) with a back-side low-emissivity coating (emis2 = 0.04–0.13)

Frame data, sourced from both the IGDB and manufacturers, covered the following 10 types: Wood high density (U=1.9 W/m²·K), Wood low density (U=1.5 W/m²·K), Wood-aluminium (U=1.19 W/m²·K), Aluminium1 (U=4 W/m²·K), Aluminium2 (U=3.2 W/m²·K), Aluminium3 (U=0.9 W/m²·K), Aluminium4 (U=0.71 W/m²·K), Vinyl1 (U=2.2 W/m²·K), Vinyl2 (U=1.8 W/m²·K), Vinyl3 (U=0.66 W/m²·K).

Seven shading control (SC) programs were applied in simulations by combining two annual schedules and two activation methods, solar radiation or indoor temperature. The two SC schedules were:

- AnnualWindowShading1: Blinds retract in winter (Jan–Mar, Nov–Dec) and inter-seasonal months (Apr, May, Sep, Oct), activating when solar radiation exceeds 200 W/m² [80].
- AnnualWindowShading2: Blinds retract in winter but activate from Apr–Oct above the same radiation threshold.

In both schedules, blinds retract at night. The optimization tool yielded solutions incorporating one of these seven SC:

- SC0: No shading; slats always retracted.
- SC1: Blinds activate per AnnualWindowShading1 when radiation exceeds 200 W/m²; slats rotate to block sunlight.
- SC2: Same as SC1, but slats remain horizontal (90°).
- SC3: Blinds activate per AnnualWindowShading2 when radiation exceeds 200 W/m²; slats rotate to block sunlight.
- SC4: Same as SC3, but slats remain horizontal (90°).
- SC5: Blinds activate when solar radiation exceeds 200 W/m² and indoor temperature surpasses 27°C; slats rotate to block sunlight.
- SC6: Same as SC5, but slats remain horizontal (90°).

During simulations, frames and blind slats solar reflectance varied simultaneously from 0.29 to 0.85 (**Table 2**), based on the Cool Roof Rating Council's paint catalog [86].

2.2. Optimizer development

The goal of an optimization algorithm is to identify the optimal solution to a function known as the fitness function. For single-objective optimization, the fitness function is generally defined by Equation 1:

$$f(X) \rightarrow y \qquad (1)$$

$$X = (x_1, x_2, \ldots, x_n) \quad \forall \, x_i \in \mathbb{S}$$

being X one solution of the searching space $\mathbb{S}$ and $y$ the associated fitness value mapped using the objective function $f$. For this article, $f$ is considered to be a continuous function $f: \mathbb{R} \rightarrow \mathbb{R}$ whose boundaries are independently defined for each of its components $x_i \subset [s_{il}, s_{ih}]$. The ultimate goal of an optimization algorithm is to find the $x^*$ that meets Equation 2:

$$:\nexists x \in \mathbb{X} : f(x) < f(x^*) \qquad (2)$$

In this research, the fitness function simultaneously addresses multiple goals, categorized as Quality and Penalty functions (see Sections 2.2.2 and 2.2.3). Each goal was defined following the method described in Section 2.2.1 aimed to leverage its importance depending solely on its numerical outputs. Thereafter, goals were combined using the weighting method outlined in Section 2.2.1.

*2.2.1 Weighting different objectives and normalization.*

In multi-objective optimization (MOO), several approaches can be employed. One common method is to transform the problem into a single-objective optimization by combining objectives using a weighted sum. Another approach involves using specialized algorithms, such as NSGA-II [87], which do not guarantee the simultaneous optimization of all objectives, as they may be conflicting. In this case, a solution is considered nondominated, Pareto efficient, or noninferior if no objective can be improved without compromising at least one other.

In this study, we combine the objective functions into a weighted sum to define the fitness function. This approach is guided by two primary considerations. First, it provides a systematic means to compare different objectives, facilitating the identification of clear preferences among nondominated solutions. For example, a solution with a heating demand of 30 kWh/m² and a cooling demand of 40 kWh/m² is not formally dominated by another solution with a heating demand of 200 kWh/m² and a cooling demand of 39 kWh/m². Nevertheless, from an architectural standpoint, the former is evidently superior. Second, as detailed in later sections, this integrated method allows the algorithm to prioritize meeting the regulatory thresholds for all objectives. Once these thresholds are achieved, the focus shifts to minimizing material usage in windows and shading elements.

In addition, to the combination of the different objectives, four penalty functions, described in Section 2.2.3, are included to ensure normative compliance. Furthermore, to make sure these objectives are on a comparable scale and that our weighting effects are not biased by being on a different scale, we apply a normalization function N(X) that brings all the scale ranges to [0,1] before combining them.

Let $f_i(X)$ be the fitness function associated to the $i$ objective and $p_j(X)$ be the penalty associated with the constraint $j$, the ultimate fitness function to optimize $F(X)$ is defined in Equation 3.

$$F(X) = \sum_i w_i * N(\alpha_i(X, f_i(X)) * f_i(X)) + \alpha_p * \sum_j w_j * p_j(X)) \qquad (3)$$

*2.2.2. Quality functions*

This section outlines the five OFs used to model the fitness function. While the primary objectives are to minimize EDh, EDc and NCT, experimentation revealed considerable variability in the solutions. This variability stemmed from certain elements that neither positively nor negatively affected energy demand. To address this issue, two new simplified cost functions were introduced for the size of glazing and fixed shading structures (overhangs and fins), which penalize the total fitness objective value proportionally to their surface areas. This approach prioritizes solutions that reduce material usage (see third objective in Section 2.1.1), as long as they achieve similar outcomes for other objectives. However, future research should focus on developing more advanced methods that incorporate realistic cost and emission factors into the optimization process. Further details on how similarity was defined are provided in Section 2.2.7.

To compute the building's energy demand, the heating and cooling intensity outputs from the Energy Plus simulation are considered. Similarly, the NCT value, which aims to minimize the annual hours of uncomfortable thermal conditions as defined by the simplified model of ASHRAE Standard 55 (2004) [88], is also derived from the Energy Plus simulation. The three quality functions values are normalized according to the acceptable values of **Table 3** and a satisfaction region ratio of 1000:1 ($\alpha_s$=1/1000) was set.

**Table 3**
Normalisation and satisfaction region values for the different quality functions.

|  | **Leon** | | | **Madrid** | | | **Sevilla** | | |
| --- | --- | --- | --- | --- | --- | --- | --- | --- | --- |
|  | Min | Satisfactory zone | Max | Min | Satisfactory zone | Max | Min | Satisfactory zone | Max |
| EDh | 0 | 46 | 70 | 0 | 30 | 70 | 0 | 12 | 70 |
| EDc | 0 | 10 | 70 | 0 | 15 | 70 | 0 | 20 | 70 |
| NCT | 0 | 438 | 700 | 0 | 438 | 700 | 0 | 438 | 700 |

*2.2.3. Penalty and satisfaction region*

A continuous fitness function, as described above, consists of an infinite set of points subjected to be inside the searching space (range of values). However, real-world problems often involve complex constraints, such as binding energy limits, which restrict the range values of many variables. These constraints can be handled in two ways: either by discarding samples once they are classified as non-compliant, or evaluating them and applying a penalty to differentiate them from non-compliant solutions. Discarding samples saves resources by avoiding unnecessary evaluations, but creates empty regions within the search space, which effectively hardens the task of exploring the boundaries of the problem, which can only be approached from the feasible side. As constraints become increasingly restrictive, the problem may become a feasibility problem, in which acceptable regions become so small that the algorithm struggles to find compliant solutions. Alternatively, evaluating and penalizing solutions can lead to higher resource usage but ensures that samples keep being generated across the entire search space. By using value-dependent penalties, whose effect increases as the constraint is further violated, additional guidance to direct the algorithm toward feasible regions is introduced. For instance, optimization algorithms that use the gradient can use penalty information to find guidance to lead the candidates to a non-penalized region. A penalized fitness function $f_p$ can be described by Equation 4:

$$f_p(X) = f(X) + \alpha_p * p(X) \tag{4}$$

where $f(X)$ is the fitness function; $\alpha_p$ is the fitness-penalty ratio, that allows to establish the ratio between the magnitude of the penalty and the magnitude of the fitness function; and $p(X)$ is the penalty associated to the solution $X$. As a general rule, $\alpha_p$ is configured so that Equation 5 is met.

$$f(X) \ll \alpha_p * p(X) \tag{5}$$

On the other hand, satisfaction regions are modifiers of the fitness function that lowers the importance of some of its components if its value is considered good according to some expert knowledge, acting in the opposite direction to penalties which increase the importance of some components. A fitness function with satisfaction regions $f_s$ can be defined as shown in Equation 6.

$$f_s(X) = \alpha(X, f(X)) * f(X) \quad (6)$$

$$\text{where } \alpha(X, f(X)) = \begin{cases} 1 \text{ if } C \\ \alpha_s \text{ if not } C \end{cases}$$

$\alpha_s$ being the desired ratio between the fitness function inside and outside the satisfaction region, and C the condition to activate the satisfaction region.

The fitness function with penalty and satisfaction $f_{sp}$ region is depicted in Equation 7:

$$f_{sp}(X) = \alpha(X, f(X)) * f(X) + \alpha_p * p(X) \quad (7)$$

The penalty part of the objective function ($\alpha_p * \sum_j w_j * p_j(X)$ in Equation 6) is calculated using a fitness-penalty ratio of 1:1000 ($\alpha_p$=1000) and the weight for each penalty is set to $w_j$=1. Also, $p_j$ is limited to positive values. Once the penalty target is reached its value remains constant at 0 (no reward for getting far from penalty's boundary).

The penalty functions are explained in detail below:

**Solar control of the envelope.** The current NEC [89] limits the solar heat gains through windows into a thermal zone by the parameter $Q_{sol,Jul}$, which is the ratio between the solar gains entering a zone in the month of July (when the mobile shading devices are activated) and its useful area. The upper acceptance limit for $Q_{sol,Jul}$ in any zone is 2 kWh/m². This penalty is calculated as shown in Equation 8. This equation generates a linear penalty whenever the threshold is exceeded and is configured to smoothly transit by ensuring $p_{solar}(2) = 0.0$.

$$p_{solar}(x) = \frac{1}{3.5-2} * (x - 2) \quad (8)$$

**Thermal transmittance of windows ($U_w$ in W/m²·K).** The thermal transmittance of a component of the building envelope (U-value) is the density of heat flow across a flat assembly, in W/m², at a temperature difference of 1 K between the surroundings of both surfaces [90]. U-value defines the ability of an element to transmit heat under steady-state conditions and it does not take into account the dynamic, time-dependent, oscillations of outdoor and indoor temperatures. For this reason, the Spanish NEC establishes upper acceptance limits for the U-value of building components depending on the winter severity of the building location. For example, for windows of buildings in Sevilla (mild winters), considering the joint effect of glazing and frame, the upper acceptance limit of U-value is 2.3 W/m²·K, while for that in León or Madrid (colder winters) is 1.8 W/m²·K.

Let $U_g$ be the glass thermal transmittance, $A_g$ be the area of the glass, $U_f$ the thermal transmittance of the frame, $A_f$ the area of the frame, $l_{g-f}$ the contact length between the frame and the glass and $\psi_{g-f}$ the linear thermal transmittance between the glass and the frame, the thermal transmittance of a window $U_w$ can be defined as follows according to EN ISO 10077 [91]:

$$U_w = \frac{U_g * A_g + U_f * A_f + l_{g-f} * \psi_{g-f}}{A_g + A_f} \qquad (9)$$

Afterwards, the penalty is calculated evaluating $U_w$ as shown in Equation 10, where $x_{lim}$ is the limit imposed by the NEC for the region under consideration.

$$p_{tt}(x) = \frac{1}{5.5 - x_{lim}} * (x - x_{lim}) \qquad (10)$$

**Heat transfer global coefficient (K in W/m²·K).** It is the mean value of the heat transfer coefficient (H) for the heat exchange surface of the envelope of a building, considering linear and punctual thermal bridging. It is expressed as:

K= [ $\Sigma_i$ A$_{x,i}$ U$_{x,i}$ + $\Sigma_k$ l$_{x,k}$ ψ$_{x,k}$ + $\Sigma_j$ x$_{x,j}$ ] / $\Sigma_x$ $\Sigma_i$ A$_{x,i}$ (11)

Where:

x refers to each element of the envelope, including those elements in contact with the ground, with the external environment, and excluding those in contact with other buildings or other adjacent spaces.

$A_{x,i}$ refers to the heat-exchange area of the element under consideration (x), in m²,

$U_{x,i}$ refers to the thermal transmittance (U-value) of the element under consideration (x), in W/m²·K,

$l_{x,k}$ is the length of the thermal bridge of the element x,

$\psi_{x,k}$ is the value of the linear thermal transmittance of the thermal bridge under consideration,

$x_{x,j}$ is the punctual thermal bridge of element x.

The upper acceptance limit of K is also regulated in the NEC, as a function of the winter severity of each specific zone and of the Compactness Index (CI) of the building: $K_{max}(Region, CI)$. CI is the ratio of the volume enclosed by the envelope to the outer surface and it is measured in m³/m². For the case study building the CI is 2.7m³/m² and the corresponding upper acceptance limits for K ($K_{max}$) are 0.59, 0.54 and 0.69 W/m²·K for Madrid, León and Sevilla respectively.

As a first approximation to the general problem of optimizing ZEB fenestration, the continuity of the insulation of the façade and the joinery was assumed, thus avoiding the appearance of significant linear thermal bridges at the contact between the opaque part of the envelope and the fenestration. The same applies to the junction between the glass and the frame in windows. Thus, the thermal bridges were neglected in the $U_w$ and K calculations, averting the limitations of Energy Plus in this respect [74],[92] and performing overly complicated pre-simulation operations. K is then calculated using the Equation 12:

$$K = \frac{\sum_i A_{x,i} \cdot U_{x,i}}{\sum_i A_i} \tag{12}$$

Finally, the value is evaluated into a linear penalty function (Equation 13) that ensures a smooth transition between climate zones:

$$p_{htc}(x) = \frac{1}{0.9 - K_{max}} * (x - K_{max}) \qquad (13)$$

**Minimum glazing area**. This penalty function ensures a baseline level of daylighting within the thermal zone, preventing unrealistic results from extremely small windows that would minimize energy demand, but fails to meet daylighting recommendations. Although in Spain, regulations do not require a minimum level of natural lighting in residential, municipal regulations of Madrid mandate that window areas in rooms must cover at least 12% of the floor area to meet hygiene standards. Then, the window to floor ratio ($R_{wf}$) was established with the acceptance limit $R_{wf,min} = 0.12$. $R_{wf}$ can be calculated as in Equation 14 and the penalty function is described in Equation 15.

$$R_{wf} = \frac{\sum window\ area}{Floor\ area} \qquad (14)$$

$$p_{rwf}(x) = \frac{1}{R_{min}} * (R_{min} - x) \qquad (15)$$

*2.2.4. Optimization algorithms*

The effectiveness of the optimization process is highly dependent on the choice of the optimization algorithm. While some algorithms may perform best on many problems, there is no guarantee that those same algorithms perform equally well on a distinct set of problems [93]. In this research a complex state-of-the-art hybrid method: SHADE+L-BFGS-B with restarts was selected. The Adaptive Differential Evolution (SHADE) algorithm is an advanced variant of the DE algorithm designed to improve convergence speed and solution accuracy for complex optimization problems. Introduced by Tanabe and Fukunaga in 2013 [23], SHADE enhances the adaptability of DE by incorporating a historical memory of successful parameter settings. This memory stores information about successful mutation and crossover parameters (mutation factor F and crossover rate CR) from previous generations. By using this historical data, SHADE dynamically adjusts these parameters to maintain a balance between exploration and exploitation, which is crucial for avoiding premature convergence and effectively navigating the search space. SHADE has been recently combined with several Local Search methods to better exploit the search space and fine-tune the solutions found by the SHADE algorithm. A relevant contribution in this sense is the SHADE-ILS algorithm, proposed in 2018 by Molina et al. [22], that combines SHADE with two local searches: The Multiple Trajectory Search (MTS) algorithm and the Limited-memory Broyden–Fletcher–Goldfarb–Shanno (L-BFGS) algorithm (a quasi-newton method with fast convergence towards the optimum and limited memory footprint). It also includes an intelligent restart mechanism to avoid premature convergence. In our experiments, this algorithm was considered, but only one of the local searches was used, as the effect of the second one did not have a significant impact on the final results.

*2.2.5. Encoding space*

The problem at hand involves both continuous variables, such as glazing and shading structures sizes, and discrete variables with a limited set of combinations (e.g., solar control types, frame materials). To address this, discrete decisions were characterized using continuous variables. For instance, glazing (the combination of two glasses and a gas chamber) was characterized by their Thermal transmittance (U), Visible transmittance (VT), and Solar heat gain coefficient (SHGC). By unifying the input variables in the continuous domain, two main steps involved in evaluating the fitness function ($f$) were identified: canonicalization ($c$) and evaluation ($e$). Given a solution $X \in \mathbb{R}$, it needs to be transformed to its canonical building $X´$ in step $c$, which generates the unique identifier for the set of solutions that ends in the exact same proposal. To calculate $X´$ from $X$, the nearest neighbor approach is followed. $X'$ is then evaluated by Energy Plus to generate the output $Y$.

$$f(X) \sim (e \cdot c)(X) \sim X \Rightarrow X' \Rightarrow Y$$

This method artificially increases the size of the input space when converting the discrete variables into the continuous domain, increasing the complexity of the problem. However, the new input space could easily be clustered into the different neighborhoods represented by its canonical building $X'$. By effectively using a cache to prevent $X'$ from being evaluated twice, the impact of this change in computational resources was reduced. Each group of variables are characterized as follows:

- Shading Control (SC) is characterized by control type, schedule type, and slat angle control type. Discrete values are approximated by the nearest feasible neighbor.
- Frame material is defined by its thermal transmittance and assigned using the nearest neighbor.
- Glazing dimensions, for which width and height values are truncated to the first decimal for faster convergence and real-world tolerance (special behavior 1).
- Wall glazing refers to the type of glazing composition, characterized by U, VT, and SHGC parameters, with values assigned to the nearest neighbor.
- Fixed shading structures were characterized by extension and depth (see Appendix C). Values are truncated to the first decimal (special behavior 1), structures are only built if values exceed 20 cm (special behavior 2). Overlaps between overhang and fin extensions are clipped to 7 cm (special behavior 3). Values are assigned to the nearest neighbor.

Tables II, III and IV of Supplementary Material show the complete coding space used for the described target in the sample building. When extending this solution to other buildings, the dimensionality of the problem is likely to be modified depending on the number of elements to optimize.

*2.2.6. Experimental scenario*

The fitness function includes several parameters that requires configuration to achieve the desired results. It was designed to have three levels of importance that must be satisfied in order to move to the lower level: penalties, fitness and satisfaction. Solutions that do not meet the normative are fully driven by the penalty functions. Solutions that can be developed but do not meet the energy efficiency target (satisfaction) are driven by three energy-efficiency functions: EDh, EDc and NCT. Finally, solutions that meet both the normative and the satisfaction targets are driven by the simplified cost function, aiming to minimize the amount of material used.

Each proposed solution is located in a zone depending on its normative and standard compliance. Hence, the proportion between the zones ($\alpha_p$ and $\alpha_s$) was the first parameter to be configured. The first zone ratio refers to the relationship between the penalty zone and the fitness zone, where solutions can be found that do not meet the regulatory requirements. In theory, these solutions can be discarded, but by measuring how far they are from meeting the targets, the optimizer can be guided towards valid solutions. When the optimizer is exploring this zone, penalties should dominate the fitness function, thus, penalty to fitness ration was set to 1000 ($\alpha_p$). It is important to consider that the output values of the different functions are normalized before weighing. Therefore, the penalties are multiple orders of magnitude bigger than the other functions. Once the regulations are met, it is energy efficiency — the main focus of this paper — the zone that led the way to optimization. These functions follow some criteria, usually defined by the standard to be met, and focused on several objectives simultaneously. However, multi-objective problems are generally highly dependent on the relationship between their variables. To address this issue, satisfaction regions were introduced, allowing certain objectives to be disabled once their targets are met, enabling a greater focus on other metrics. The third level of importance of the fitness function is the satisfaction region, whose ratio was set to 1000 ($\alpha_s$), equal to the penalty-fitness ratio, which is high enough to make the other fitness functions lead the optimization when this target is satisfied. This level is reached when the solution is below the values defined in Section 2.2.2.

When all, EDh, EDc and NCT were met, the driver of the problem was the simplified cost (see Section 2.1.1), which measures the amount of material used and is generally correlated with the emissions generated to produce the resource. The equivalence between 1 kWh/m² heating and 1 kWh/m² cooling was also established. To achieve this, it was necessary to ensure that the normalization range (max-min), which in this case is 50 kWh/m², is identical for both variables. For instance, the maximum EDh value considered for Madrid is 80 kWh/m² because the satisfaction value is reached at 30 kWh/m². This approach also establishes that 1 kWh/m² over the limit is equivalent for both variables at the same time, avoiding the artefacts that can occur due to target inequality. In addition, the minimum value was reduced by 5 kWh/m² to prevent cases where the value near the target approaches zero. This effectively prevented lower order functions (simplified cost) from driving optimization before the function's targets are reached (with no margin all decimals must be exceeded before the target is reached and cost functions become comparable at some point). For the NCT case, a min-max range of 500 hours was set, which seems reasonable given the preliminary results of the problem. In all cases, the satisfaction value was set at 438 hours, so the minimum and maximum for normalization were set at 388 and 938 hours respectively. Similarly, fixed shading and window simplification costs are calculated by setting values to the maximum and minimum allowed extensions (search space permitted values, not penalization values). The weights shown in **Table 4** are calculated according to real-world parameters using Equation 17, to minimize the influence of some non-domain parameters such as the min-max range in the fitness function, which only affect normalized values.

$$w_v = \alpha_v * \frac{R_v}{A * R_E} \tag{17}$$

Where:

$w_v$ is the resulting weight of the variable $v$;

$\alpha_v$ is the real-world relation between a unit of $v$ and 1 kWh;

$R_v$ is min-max range expressed in the units of $v$;

A refers to the area of the thermal zone (60 m²);

$R_E$ is the range size of the heating and cooling (50 kWh/m²)

For example, for NCT the following relation is used:

$\alpha_{NCT} = 6\ hours/kWh$ , which generated a weight $W_{NCT} = 1.0$.

**Table 4**
Parameters for the different quality functions.

| Metric | Units | Weight | α | Minimum | Maximum | Satisfaction Value |
|---|---|---|---|---|---|---|
| | | | | | León / Madrid / Sevilla | |
| EDh | kWh/m² | 1 | N/A | 41/25/7 | 96/80/62 | 46/30/12 |
| EDc | kWh/m² | 1 | N/A | 5/10/15 | 70/65/60 | 20/15/10 |
| NCT | Hours | 1 | 6 | 388 | 938 | 438 |
| Fixed shading cost | m² | 0.295 | 20 | 0 | 44.27 | -∞ |
| Window cost | m² | 0.291 | 60 | 3 | 17.57 | -∞ |

*2.2.7. Hyperparameter selection*

The configuration of the hyperparameters of the optimization algorithms has a major impact on their overall performance. However, due to the large number of possible parameters, a full exploration was not possible. Therefore, where available, the recommended values from the literature were chosen. Alternatively, some parameters were explored when this information was not available. For the SHADE algorithm, the possible values of H were set to either 50 or 100, with 50 being chosen. The size of the population varied between 15, 30 and 60, with 60 being chosen (also for the DE). Moreover, the eps parameter of L-BFGS-B was set to 0.1 after testing the values 0.1 and 1e-8. The restart criteria are based on a combination of two parameters: the restart threshold (*thres*) and the number of steps with no improvement (*NI*). The former controls the percentage difference between the best sample in the input population and the best sample in the new population at the end of this algorithm execution (one algorithm gives way to the other) necessary to consider the step as an improvement. The latter controls the number of consecutive executions allowed without improvement before the restart is triggered. NI was set to 3, as proposed in [22] for a similar hybrid algorithm. However, for this research, the parameter *thres* was set to 1e-4 in order to let the algorithm converge and only trigger a restart when it is completely stagnated. It is worth considering that the chosen coding space is likely to produce similar buildings when converging, as the solution evaluated is its nearest neighbor. Therefore, 1e-4 should be sufficient to detect stagnation. Higher-order values were also tried during the problem definition, but they caused an early restart. The following **Table 5** gives a summary of the parameters tested and selected. Further research is encouraged to find the best performing parameters for this type of problem.

**Table 5**
Analysed hyperparameter values and selected values

| Algorithm | Hyperparameter | Tested values | Selected value |
|---|---|---|---|
| DE | population size | 60 | 60 |
|  | F | 0.8 | 0.8 |
|  | CR | 0.5 | 0.5 |
| SHADE | pop_size | 15, 30, 60 | 60 |
|  | H | 50, 100 | 50 |
| L-BFGS-B | eps | 0.1, 1e-8 | 1.00E-08 |
| Restart | thres | 1.00E-04 | 1.00E-04 |
|  | NI | 3 | 3 |

## 3. Case study and energy model set-up

The method was tested in a two-bedroom, 60 m² apartment on the fourth floor of a multi-family block, one of Spain's most common building typologies [59]. With façades facing north, south, and west (**Fig.** 5) its geometry was modeled in Design Builder V6.1.8.021 and exported to EnergyPlus 8.9.0. Simulations were conducted for León, Madrid, and Sevilla to capture Spain's climate variability. Spain's climate classification [94], defines six winter severity levels (α, A, B, C, D, E) and four summer severity levels (1–4). León (E1) has the coldest winters and mildest summers, Madrid (D3) has severe winters and summers, and Sevilla (B4) has mild winters and the warmest summers (**Fig.** 6). Hourly TY weather data [77] in epw format [78] were used in the simulations.

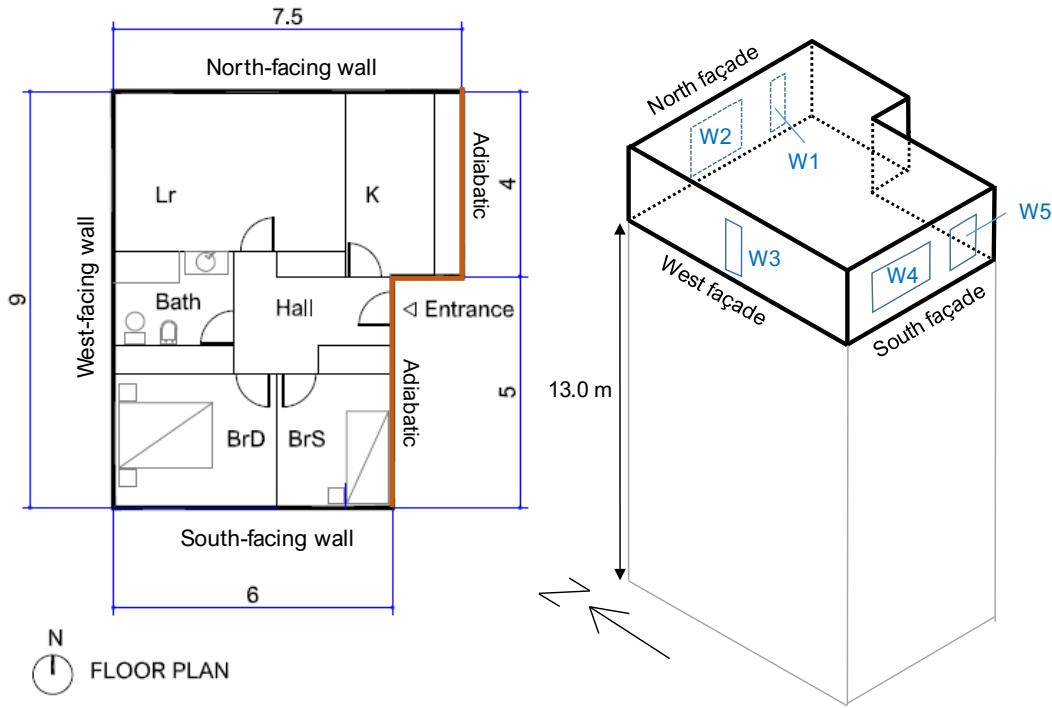

**Fig. 5.** The left side displays the floor plan of the fourth-floor corner apartment used for optimization, while the right side presents a 3D view, highlighting the windows subject to optimization. Blue lines indicate dimensions in meters.

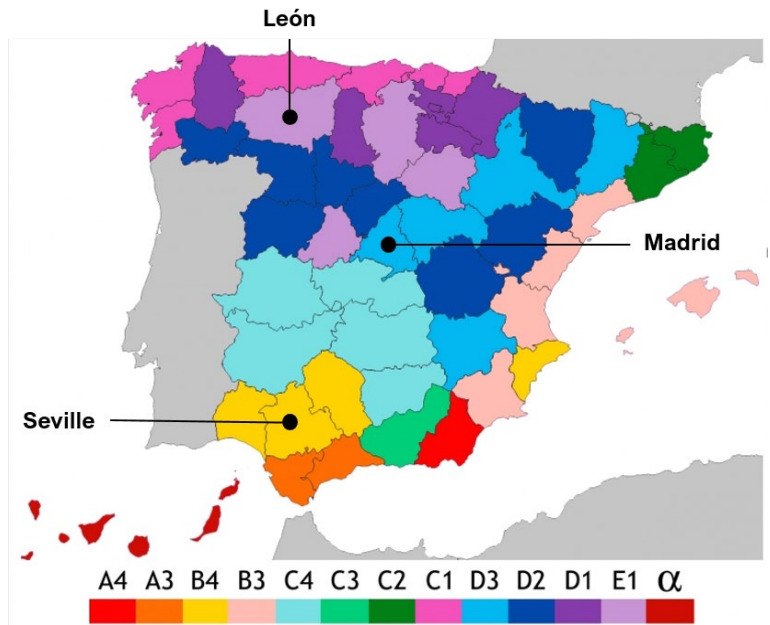

**Fig. 6**. Map of Spain's climate zones with selected cities marked. Prepared by the authors based on NEC [94]

# 4. Analysis of results and discussion

Two types of results are analysed in this section: Optimization process (Section 4.1) and Optimized fenestration solutions (Section 4.2)

4.1. Optimization process

In this section two analyses were conducted: a comparison against one of the reference algorithms of the literature and a convergence and robustness analysis of the best solutions.

*4.1.1. Algorithm comparison*

For comparing the performance of the proposed algorithm, a Genetic Algorithm with the same population and a typical configuration used in continuous domains for its parameters (**Table 6**) was executed across locations.

**Table 6**
Parameters of the Genetic Algorithm

| Parameter | Value |
|---|---|
| Population size | 60 |
| Crossover operator | BLX-alpha (alpha=0.5) |
| Crossover probability | 90% |
| Mutation operator | Gaussian mutation |
| Mutation probability | 1% |
| Selection operator | tournament 2 |

**Fig.** 7 depicts the evolution of the distribution of the best results obtained through the different 15 executions in the cities of León and Madrid. A similar result was obtained for the city of Sevilla and can be consulted on the Supplementary Material of the paper (**Fig.** A). As shown on the image, the SHADE algorithm is able to obtain not only the best median result but also it is more robust throughout all the executions converging to solutions with a similar fitness. On the contrary, the GA has a larger variance, especially in the city of León, being less robust and also converging faster and to an inferior solution.

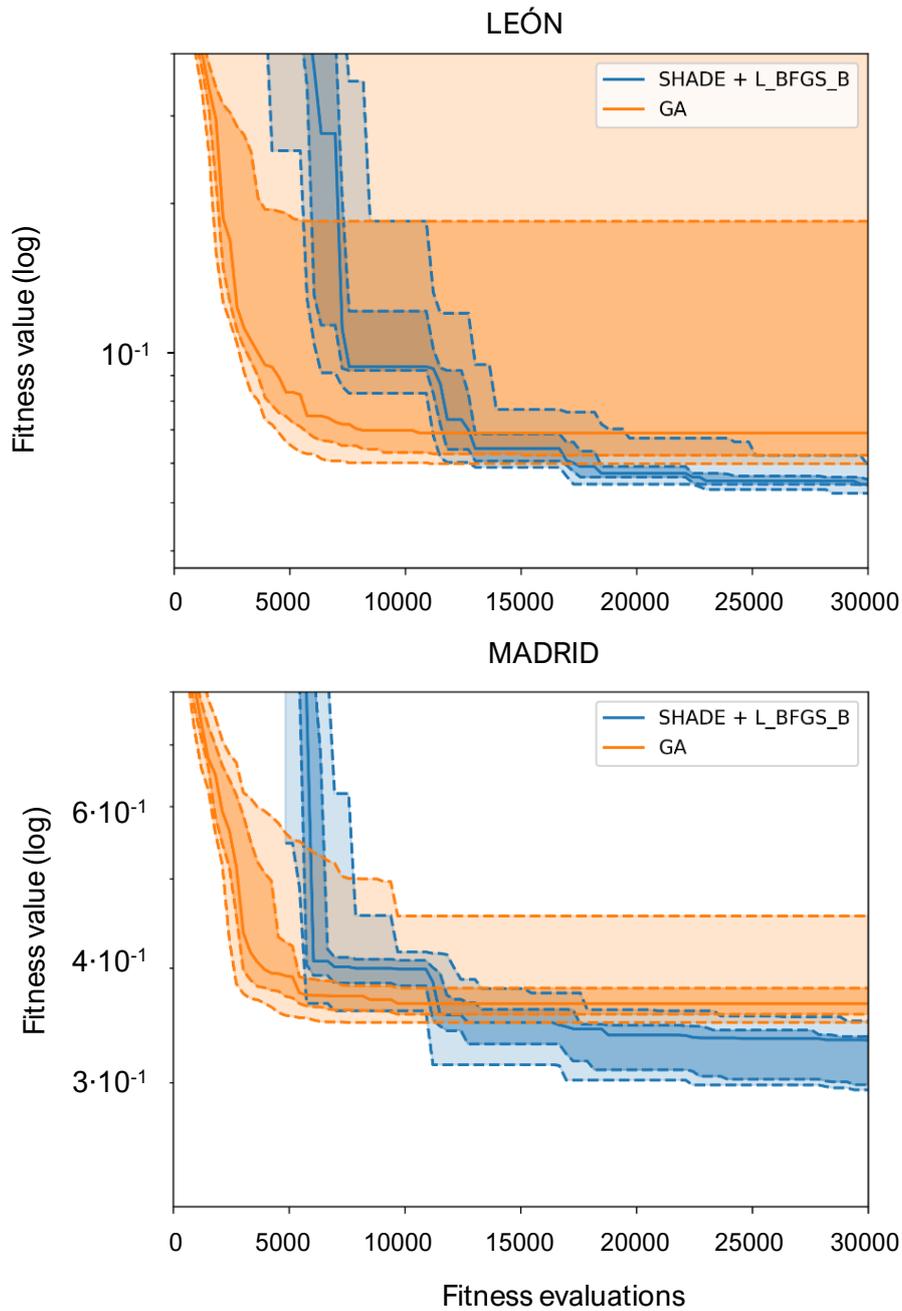

**Fig. 7.** Convergence plot for the cities of Leon and Madrid of the SHADE+L_BFGS_B algorithms vs GA. Y-axis represents the best fitness (in logarithmic scale) whereas X-axis represents the number of fitness evaluations. For each algorithm, the median value of the best results of the 15 executions is depicted by a solid line whereas the different colour regions delimited by the dashed lines represent the best 25%-75% and 5%-95% solutions.

These results were validated using a non-parametrical Wilcoxon test for the three cities. The results, presented on **Table 7** indicate that the SHADE+L-BGFGS_B algorithm significantly outperforms the GA in each city. Furthermore, after applying the Bonferroni family-wise error correction, the SHADE+L-BFGS_B algorithm remains superior across all cities, as the p-values are less than 0.0033.

**Table 7**
Statistical validation for the best solutions on the historical weather projection (*)

| SHADE-L-BGFS vs. GA | León | Madrid | Seville |
|---|---|---|---|
| Wilcoxon p-value | 3.051E-05 √ | 9.1552E-05 √ | 3.0517E-05 √ |

(*) √ means that there are statistical differences with $\alpha=0.01$

With respect to the computational time of both algorithms, each EnergyPlus simulation takes approximately 5–10 seconds. As a result, EnergyPlus simulations account for approximately 99.9% of the total execution time, rendering any differences in execution time between the two algorithms negligible.

*4.1.2. Robustness analysis*

To assess robustness, the variables from the set of the best solution for each of the 15 executions were analyzed. **Fig.** 8 illustrates the distribution of values for each variable in the best solutions for the city of Madrid. Corresponding distributions for the other cities are provided in the Supplementary Material (Figs. B and C). These results demonstrate that the SHADE+L-BFGS-B algorithm exhibits strong robustness for this problem, as the variables in the solutions obtained from each execution consistently converge to identical or highly similar values for the majority of variables.

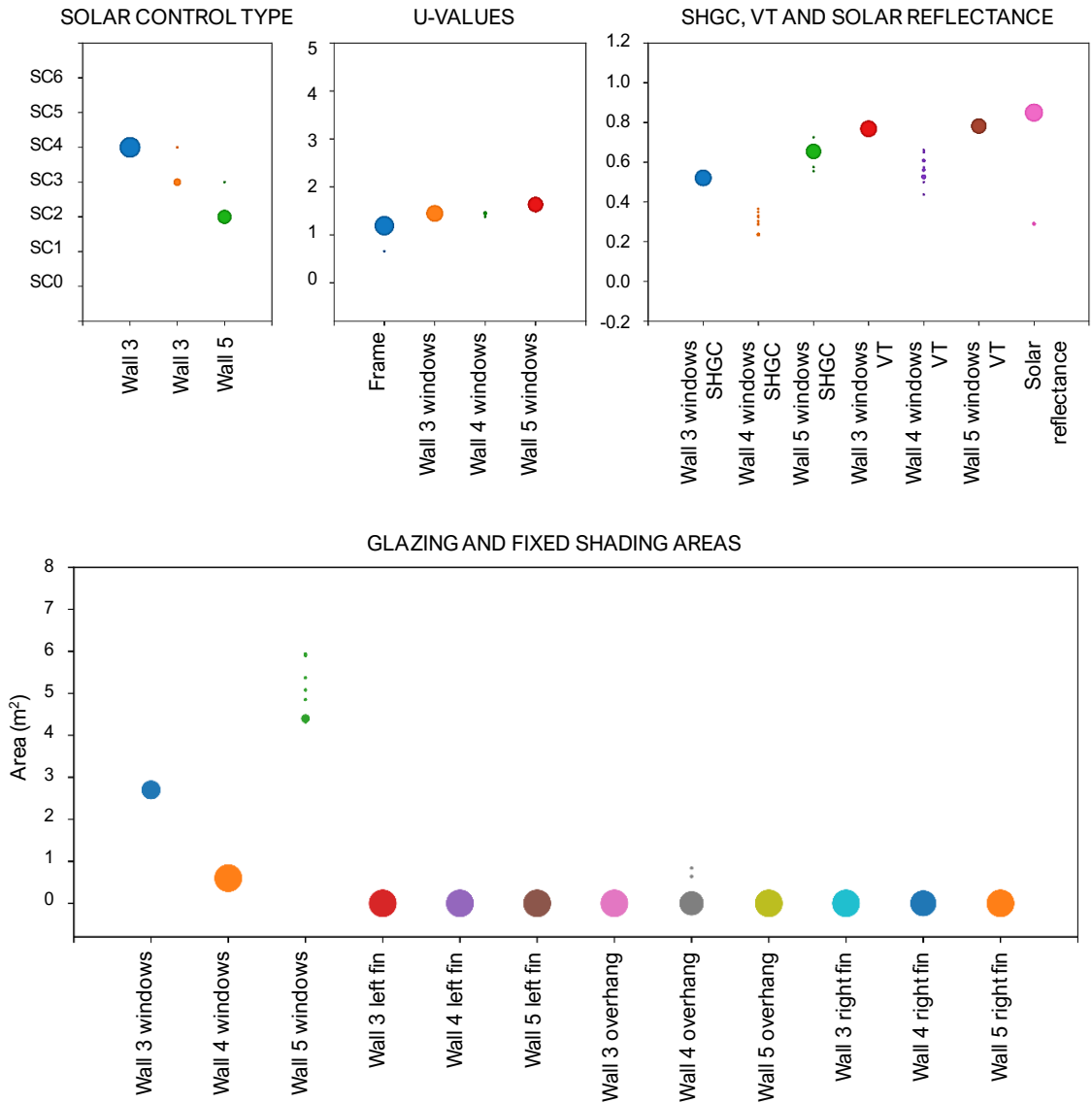

**Fig. 8**. Convergence analysis of the optimized variables across the best solutions obtained by the SHADE+L-BFGS-G algorithm from 15 executions for the city of Madrid. The X-axis represents the optimized variables, while the Y-axis indicates their corresponding values. The circle size reflects the number of executions whose best solutions share the same value for a given variable; larger diameters indicate a higher frequency of identical values across executions.

4.2. Optimized solutions

The optimization tool provided 150 fenestration solutions per location and window shading scenario (S1 where both fixed and movable shading were available and S2, where only fixed shading was available). Each solution was assigned a unique numerical identifier. Notably, none of the S1 solutions included overhangs or fins, highlighting the superior efficiency of movable shading. Three types of analysis were conducted on the optimized solutions:

1. Performance of the best solution per location and scenario to assess the maximum achievable performance when optimizing all three objectives simultaneously.
2. Variability of the Objective Functions values across the 150 solutions to evaluate the results diversity.
3. Frequency of variables values to identify common fenestration features, providing insights for future ZEBs design.

*4.2.1. Performance of the best solutions*

The best solution per location and shading scenario was selected based on the following criteria: (1) If both EDh and EDc are within the acceptance limits, the solution with the lowest combined demand was chosen, (2) If one energy demand exceeds the limit, the solution with the lowest non-compliance indicator was selected. NCT values were not considered, potentially leading to suboptimal performance in this index, as will be shown below. **Fig.** 9 illustrates the glazing composition code for two solutions, highlighting the tool's capability to provide detailed information directly applicable to architectural prescriptions. **Table 8** presents the genome of the best solution for Madrid as an example.

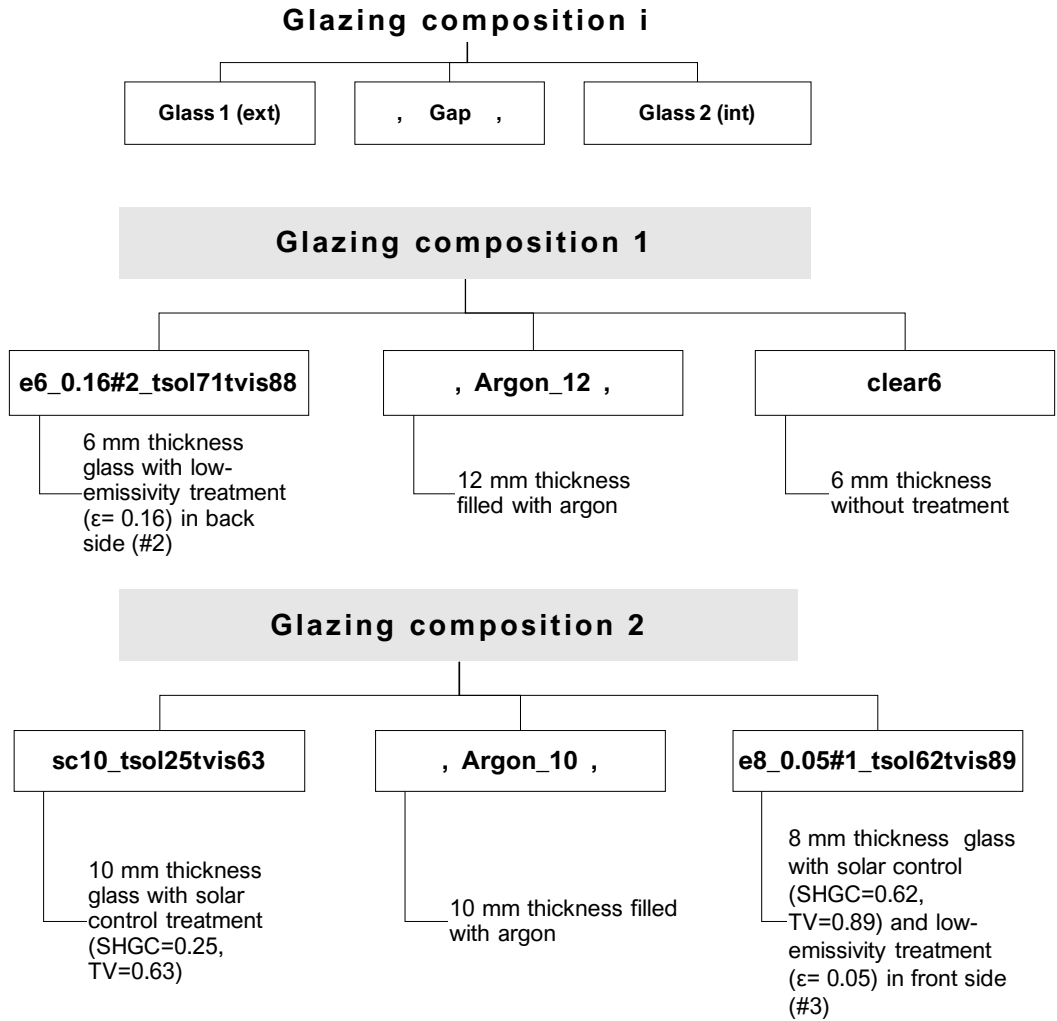

**Fig. 9.** In this figure, the code detailing the glazing composition for each façade is depicted, along with two examples of solutions generated by the optimization tool. These examples demonstrate the tool's ability to offer precise information that can be directly utilized in architectural specifications.

**Table 8**
Example of the best solution for Madrid (Genome 14659TY)

|  | Objective and penalty functions | Units | Value |
|---|---|---|---|
| *Building level* | ED_Heating | kWh/m$^2$·year | 30.0 |
|  | ED_Cooling | kWh/m$^2$·year | 19.3 |
|  | ED_Heating+Cooling | kWh/m$^2$·year | 49.3 |
|  | NCT | h | 147 |
|  | WWR | % | 15 |
|  | K | W/m$^2$·K·year | 0.48 |
|  | Q$_{sol,Jul}$ | kWh/m$^2$ | 1.70 |
|  | Colour for frames and blind slats (solar reflectance coefficient) |  | Bright White / 50 ULTRA PURE WHITE, Bright White / WT3500(RC) (0.85) |
| *Component level* | **Variable** | **Units** | **Value** |
| North façade Kitchen Window1 | Width | m | 0.6 |
|  | Height | m | 1.0 |
|  | Window1_Area | m$^2$ | 0.6 |
|  | Glazing Composition |  | clear10,Argon_16,e8_0.05#1_tsol62tvis89 |
|  | Window U-value | W/m$^2$·K | 1.8 |
|  | Frame Material |  | FrameWoodAlum_Class4 |
|  | Shading Control |  | SC3 |
| North façade Living room Window2 | Width | m | 1.5 |
|  | Height | m | 1.4 |
|  | Window2_Area | m$^2$ | 2.1 |
|  | Glazing Composition |  | clear10,Argon_16,e8_0.05#1_tsol62tvis89 |
|  | Window U-value | W/m$^2$·K | 1.6 |
|  | Frame Material |  | FrameWoodAlum_Class4 |
|  | Shading Control |  | SC3 |
| West façade Bathroom Window3 | Width | m | 0.6 |
|  | Height | m | 1.0 |
|  | Window 3_Area | m$^2$ | 0.6 |
|  | Glazing Composition |  | sc10_tsol25tvis63,Argon_10,e8_0.05#1_tsol62tvis89 |
|  | Window U-value | W/m$^2$·K | 1.7 |
|  | Frame Material |  | FrameWoodAlum_Class4 |
|  | Shading Control |  | SC4 |
| South façade Single Bedroom Window 4 | Width | m | 1.5 |
|  | Height | m | 1.3 |
|  | Window4_Area | m$^2$ | 2.0 |
|  | Glazing Composition |  | e6_0.16#2_tsol71tvis88,Argon_12,clear6 |
|  | Window U-value | W/m$^2$·K | 1.8 |
|  | Frame Material |  | FrameWoodAlum_Class4 |
|  | Shading Control |  | SC4 |
| South façade Double Bedroom Window 5 | Width | m | 2.7 |
|  | Height | m | 1.5 |
|  | Window5_Area | m$^2$ | 4.1 |
|  | Glazing Composition |  | e6_0.16#2_tsol71tvis88,Argon_12,clear6 |
|  | Window U-value | W/m$^2$·K | 1.8 |
|  | Frame Material |  | FrameWoodAlum_Class4 |
|  | Shading Control |  | SC4 |

**Fig. 10** shows the EDh and EDc values of for the best solution at each location under the two scenarios. When movable shading is unavailable (S2), heating demand increased by 3, 24 and 54% in León, Madrid and Sevilla respectively, compared to the scenario where it can be used (S1). The corresponding increase in cooling demand was 15, 23 and 10 %. In León, the best solution met both heating and cooling limits under S1. In Sevilla, where winters are mild, heating demand stayed within the acceptable range, but cooling demand exceeded the limit in both scenarios. In Madrid, heating demand met the limit only under S1, while cooling exceeded the threshold in both scenarios. **Fig. 11** illustrates the impact of shading on the thermal comfort indicator (NCT). The absence of movable shading (S2) proved advantageous in Madrid, reducing the NCT by 31.4% compared to S1. In León, the correspondent NCT increased slightly by just 0.6. In Sevilla, neither scenario met the required NCT levels, indicating the difficulty of balancing strict cooling demands and thermal comfort in very warm climates.

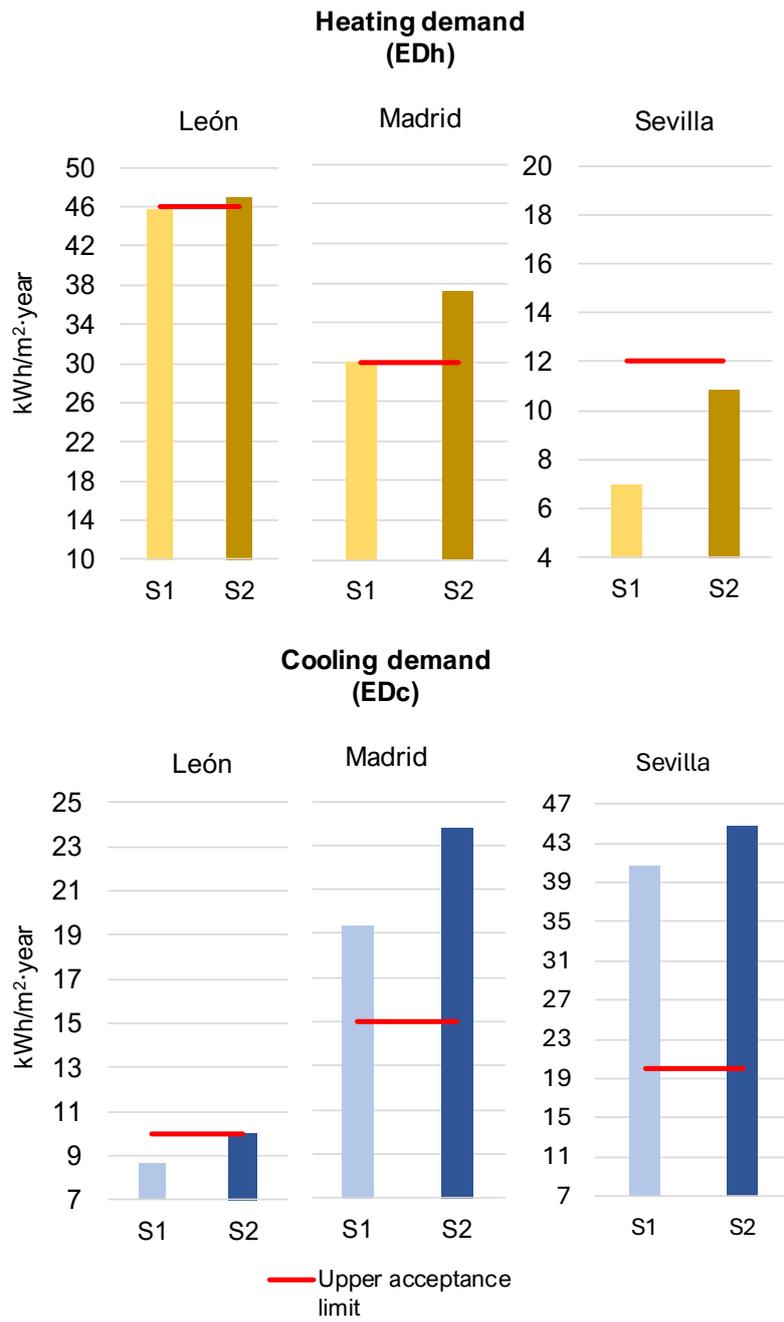

**Fig. 10**. Energy demand for heating and cooling of the best solutions for the three locations and the two design scenarios (S1 and S2)

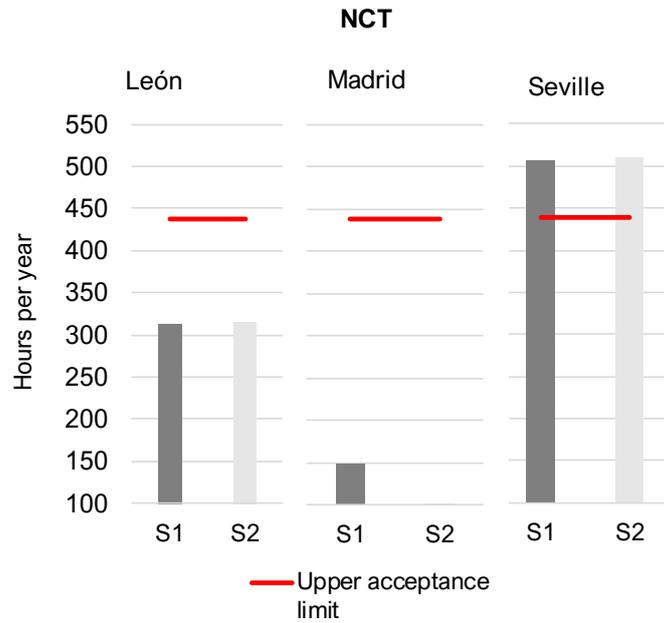

**Fig. 11**. Not comfort time (NCT) for the optimal solutions of the three locations, under S1 and S2 design scenarios.

*4.2.2. Variability of the Quality Functions values*

**Fig. 12** shows that in León and Madrid, all solutions met the upper heating demand limit under S1. However, under S2, when movable shading was unavailable, results deviated significantly. In Sevilla, all solutions remained within the heating demand limit in both scenarios. However, dispersion was higher in S1 (σ = 1.01) than in S2 (σ = 0.27). Additionally, 75% of S1 solutions required less than 7 kWh/m²·year, whereas in S2, 75% exceeded 10.5 kWh/m²·year.

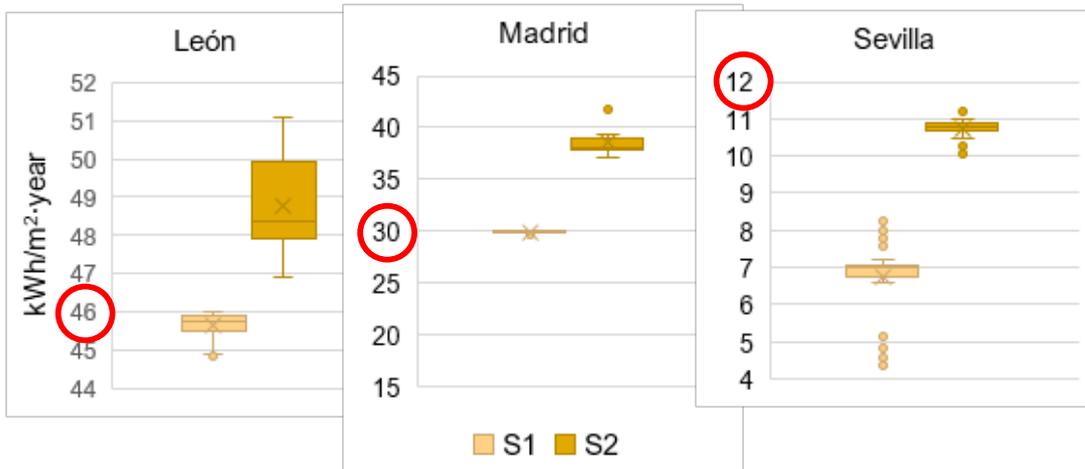

**Fig. 12**. Comparison of variability in heating demand under S1 and S2 scenarios at the three locations. The upper acceptance limit for each location is rounded by a red circle.

For cooling demand, only León's solutions met the upper limit, with a higher variability in S1 (σ = 0.33) than in S2 (σ =0.08). Solutions for Madrid and Sevilla deviated significantly in both S1 and S2 (**Fig. 13**). While solution variability was higher under S1 than under S2, lower EDc solutions were more common in S1 for all locations. These results highlight the challenge of achieving low cooling demand in regions with extreme summer conditions, like Sevilla, or even moderately severe areas like Madrid.

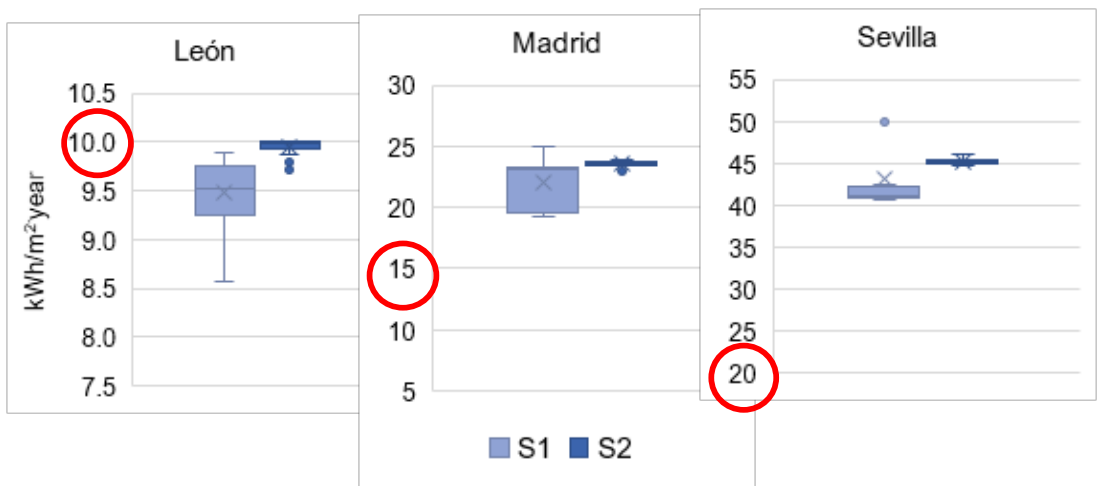

**Fig. 13**. Comparison of variability in cooling demand (EDc) under S1 and S2 scenarios at the three locations. The upper acceptance level for each location is rounded by a red circle.

For the NCT index (**Fig. 14**), all solutions in León and Madrid met the acceptance condition under both S1 and S2. In León, solutions with movable shading significantly reduced discomfort hours compared to those without it, with relatively similar variability (σ = 3.10 for S1 and σ = 5.70 for S2). In Madrid, NCT values showed more variability under S1 (σ = 28.27) than S2 (σ = 5.44), but the top 25% of S1 solutions outperformed those under S2. In Sevilla, only 30 out of 150 solutions under S1 kept discomfort hours below the upper limit (σ = 26.54), and no solution met the NCT limit under S2 (σ = 3.59). As noted in Section 4.2.1, the lowest energy demand solutions in Sevilla may fail to meet the NCT upper limit.

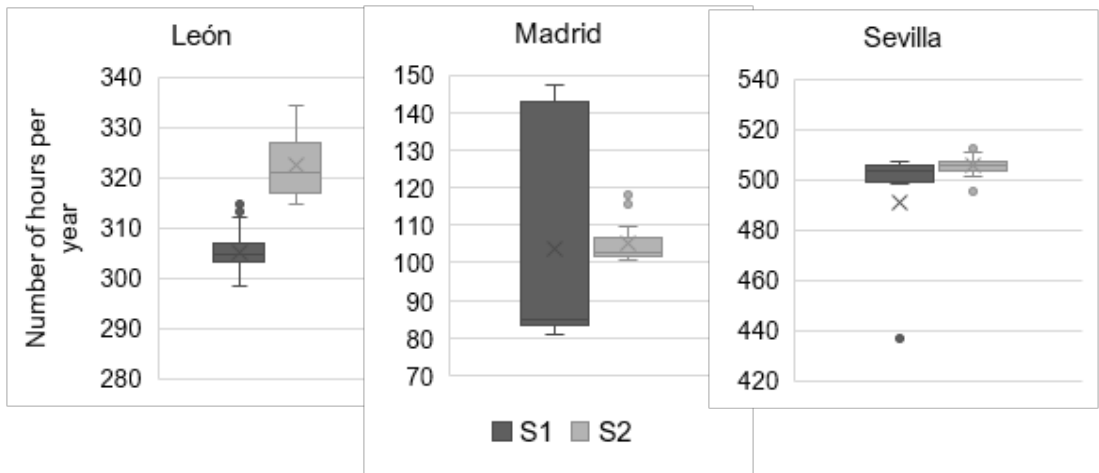

**Fig. 14**. Comparison of NCT variability under S1 and S2 scenarios at the three locations shows that the acceptable upper threshold of 438 discomfort hours per year is met in León and Madrid, but not in Sevilla.

Automated movable shading solutions consistently delivered the best performance across all three indicators. However, the shading system's importance varied depending on the climate. In warm climates like Sevilla, meeting the EDh limit was straightforward regardless of shading. However, in temperate or cold climates like Madrid and León, the type and control of shading were crucial. Automated movable shading solutions consistently met the EDh limit, whereas those with fixed shading did not always achieved compliance. In cities with severe summers (Madrid and Sevilla), EDc and NCT values showed greater dispersion with automated shading (S1) than without it (S2). In these cities, EDc σ-values were 6 and 12 times higher than in León, indicating that the impact of automated shading control increased with summer severity.

*4.2.3. Frequency of variable values*

This section analyzes the frequency of variable values that were identified as optimal solutions by the optimization tool, with the goal of identifying common fenestration attributes that could serve as benchmarks for ZEB design.

When movable and fixed shading were allowed (S1), no solutions included overhangs or fins, indicating that movable shading was more efficient. In S2, only fixed shading was available, leading to worse performance than S1, except for the NCT index in Madrid. Based on these results, key variables studied under S1 included: glazing area, composition, frame type, solar reflectance of frames and blinds, and blind control system. Under S2, overhang and fin depth values were also examined.

**Glazing area**. The glazing area of north and west facing façades remained at the minimum allowed size (Rules R7 and R8, [Appendix B](#)) regardless of location or shading scenario. In contrast, south-facing glazing area varied significantly, as shown in **Fig. 15**. In León (harsh winters), south-facing windows (W4 and W5, see Fig. 5) were larger without movable shading (S2) than with it (S1). Conversely, in warmer climates (Madrid, Sevilla), S1 enabled larger south-facing windows than in S2. In Sevilla, W4 expanded from 1 m² (S2) to 1.8 m² (S1), and W5 grew from 1.8 m² to 3.4 m², highlighting a strong link between shading systems and south-facing window size in Iberian ZEBs.

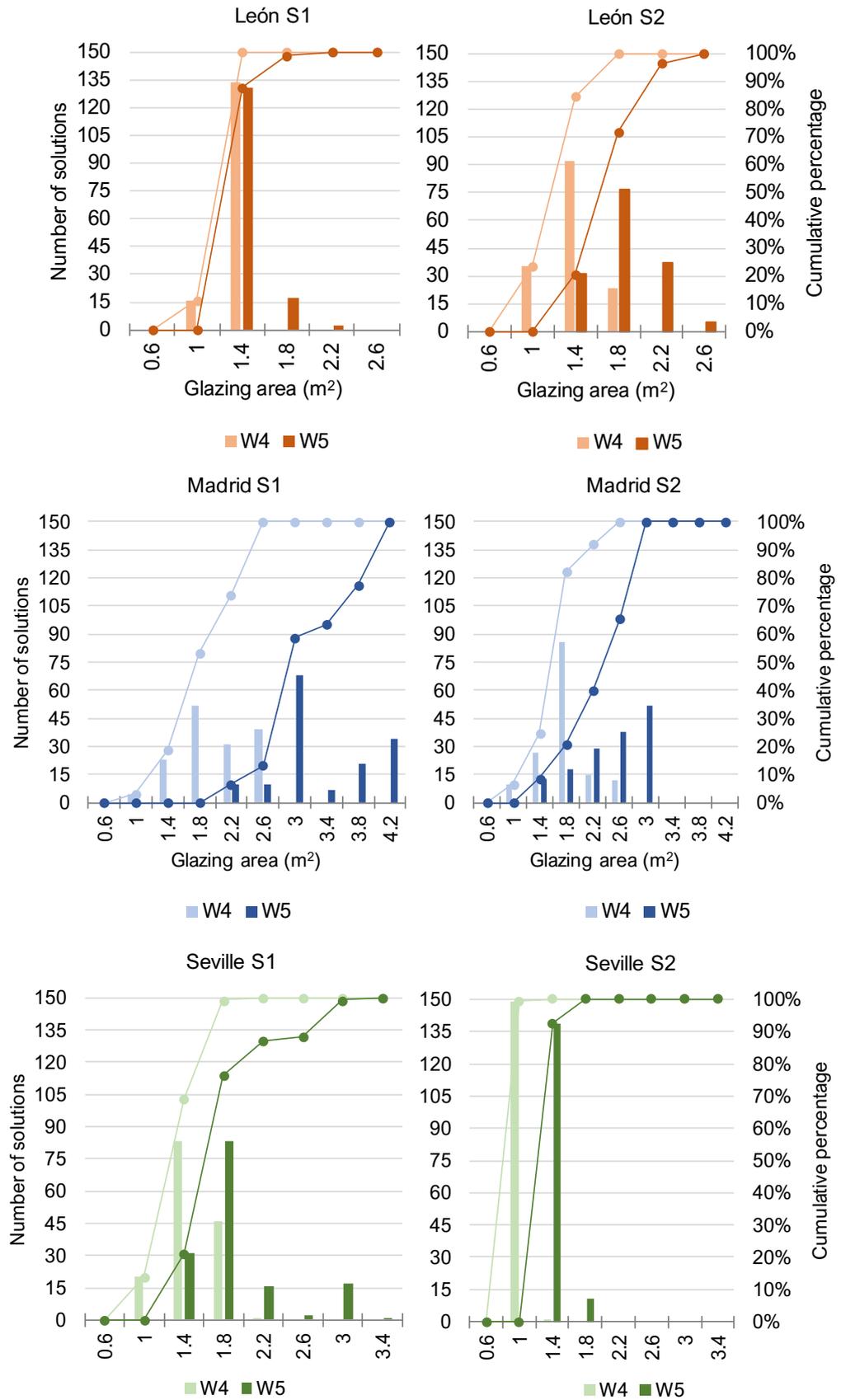

**Fig. 15**. Frequency analysis of the glazing area of windows W4 and W5 on the south facing façade for the three locations and the two design scenarios.

**Glazing composition.** Variants were analyzed by façade orientation, location, and shading scenario. Results show that western façades exhibited more variants (11–35) than other orientations (3–10) across locations and scenarios and that air was excluded as a chamber gas in all cases, with 12 mm and 16 mm gaps most commonly used. Low-emissivity treatment was consistently applied to the front of glass 2 (#3, **Fig. 2**) on north-facing façades, with very low emissivity coefficients prevailing (0.05-0.07). When only fixed shading was used (S2), all west-facing solutions always used solar control on glass 1, regardless of location. In S1, on the same orientation for León and Sevilla, options like "Clear Glass-Argon-Low-E Glass" and "Low-E Glass-Argon-Clear Glass" were added, highlighting the necessity of solar control to mitigate radiation where fixed shading alone was insufficient. In S2, the solar control glass required lower SHGC coefficients (0.23-0.37) compared to when automated movable shading was used (S1) (0.42-0.75). In S1 scenario, all locations used a single composition for south-facing glazing: "Low-emissivity Glass-Argon-Clear Glass". **Fig. 16** shows an example of frequency of glazing composition solutions for south facing façades.

**Solar reflectance of frames and blinds.** Extreme values (0.85 and 0.29) were most common, especially in S1. Only in Sevilla (S1) did slightly more variants appear. Since reflectance was uniform across all windows in the optimization, consistent with real construction practices, the low variability remains unexplained, making these results inconclusive.

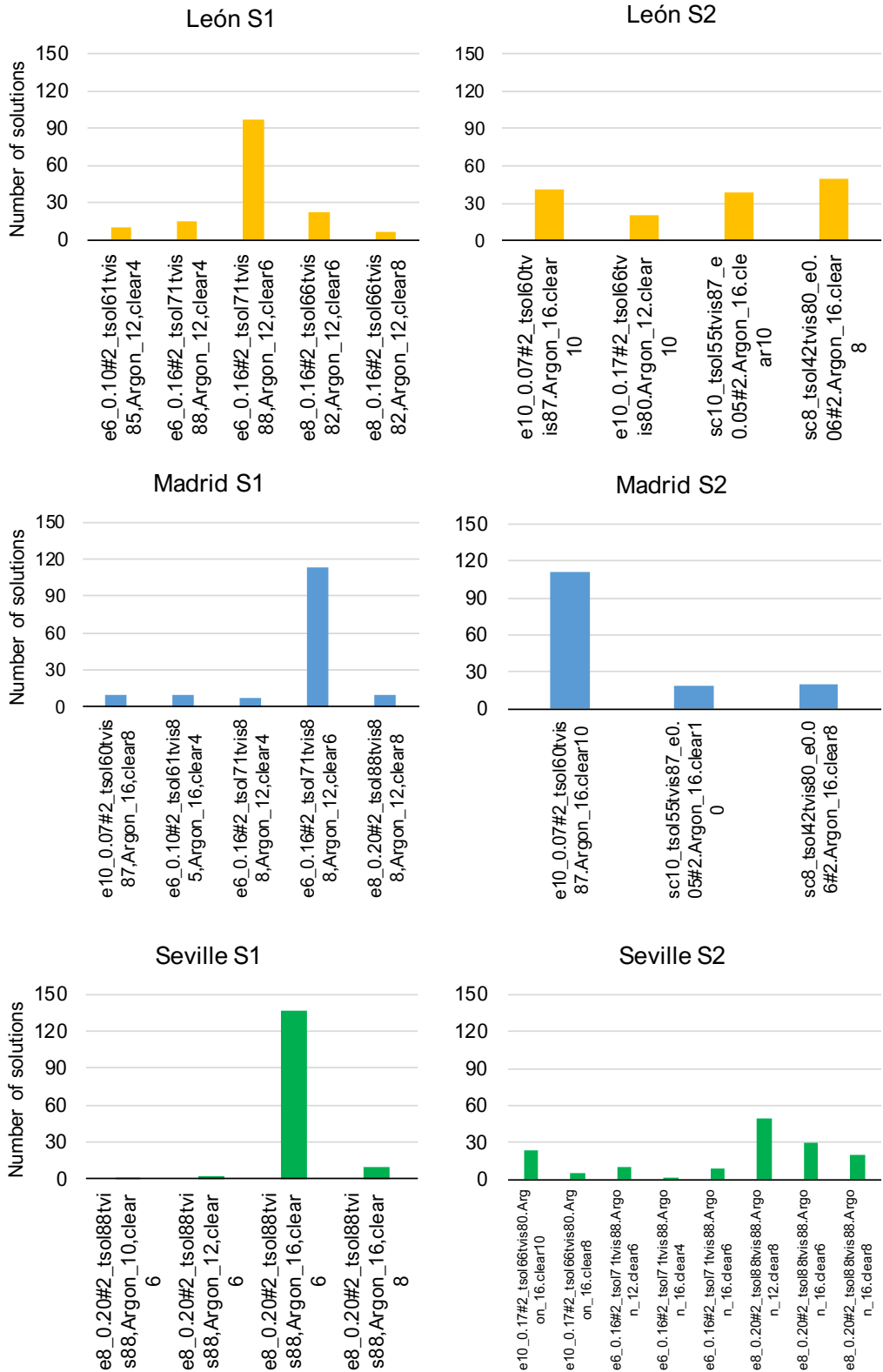

**Fig. 16**. Frequency charts of glazing composition solutions for south facing façades in León, Madrid and Sevilla, considering both shading scenarios.

**Frame type**. Of the 10 available frame types, the optimizer selected only 4 across all locations and shading scenarios, with U-values ranged from 0.66 to 1.19 W/m²·K. Frames with U-values above 1.19 were excluded. In S1, for León and Madrid, two types, WoodAlum (1.19 W/m²·K) and Vinyl3 (0.66 W/m²·K), prevailed, indicating that lower U-value frames weren't always needed with automated shading. In S2, higher-performance frames (U-value ranging from 0.66 to 0.71 W/m²·K) were more common in these locations. In Sevilla, all solutions included low or very low U-value frames (0.70–0.66), confirming the importance of this parameter in warm climates.

**Solar control systems.** The greatest variety of control programs was observed on the north-facing façade, while the least was on the south-facing façade across all locations (see Fig. D in Supplementary Material). In León, SC5 (temperature-based blind activation) was common on the north and west, while SC1 and SC2 (solar radiation-based) dominated the south, keeping blinds retracted in cold months to reduce heating loads. In Madrid, solar radiation-based programs were more common, with SC4 in the north, SC3 in the west, and SC2 in the south. In Sevilla, SC0 (blinds always retracted) was common in the north, while SC4, triggered by solar radiation and available during inter-seasonal periods, was prevalent in the west and south, addressing thermal load increases from solar gains in March, April, and October.

**Overhangs and fins in design scenario S2.** In S2, the optimizer adjusted overhang and fin depths (ranging from 0.20 to 1.50 meters; see Appendix C and **Table 2**) based on façade orientation. For north-facing façades, no solutions included overhangs or fins in Sevilla, nor overhangs in Madrid; in León, only 4 out of 150 solutions featured a minimum overhang depth of 0.20 meters. To reduce afternoon cooling loads, the optimizer recommended 83 solutions in León and 150 in Madrid with right fins on north façades, ranging from 0.2 to 0.5 meters in depth. Conversely, overhangs were prevalent on west-facing façades, appearing in 99% of León, 100% of Madrid, and 70% of Sevilla's 150 best solutions (**Fig. 17**). In Sevilla, 92% had depths under 1.0 meter, while most in León and Madrid exceeded this depth. Side fins were rare on west façades, appearing in only 4 of 150 solutions in Madrid (up to 0.40 m deep) and 12 in León (up to 0.80 m). None were present in Sevilla. On south-facing façades, overhangs appeared in all solutions but were smaller, reaching up to 0.4 m in León and Sevilla and 0.3 m in Madrid (**Fig. 17**). Side fins were absent in Madrid and Sevilla, while León had 6 solutions with left fins and 4 with right fins.

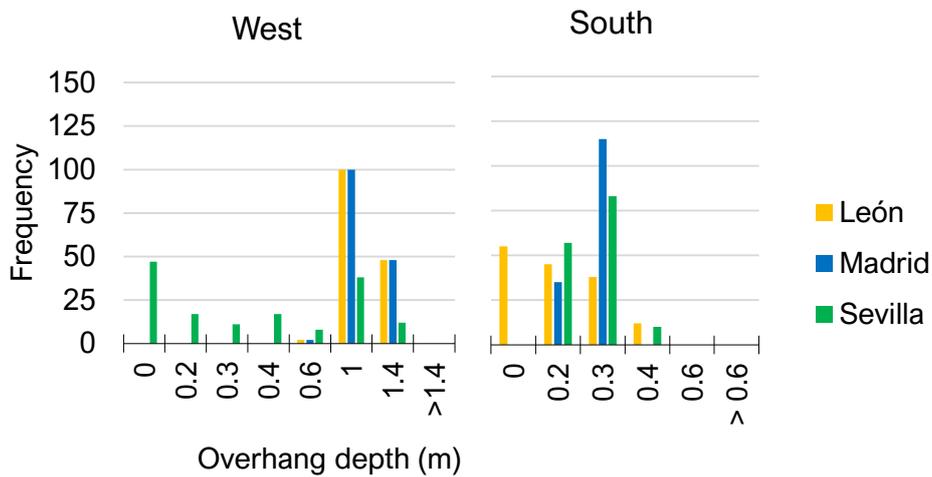

**Fig. 17**. Frequency charts depicting overhang depth in west and south facing façades across different locations.

## 5. Conclusions

A new optimization method has been developed to assist architects and engineers in the early-stage design of Zero-Emissions Buildings (ZEBs), focusing on fenestration, the most thermally inefficient and costly component of the building envelope. Nineteen fenestration variables, over which architects have design flexibility, were optimized to reduce heating and cooling energy demand, and thermal discomfort hours, while regulated variables were fixed during the optimization. The method was tested in a residential building across three Mediterranean climate locations. Using the advanced hybrid method, SHADE+L-BFGS-B with restarts, 150 optimized solutions were selected for analysis in each climatic zone. Two solar shading scenarios for windows were assessed: S1, allowing both automated movable and fixed shading systems, and S2, permitting only fixed shading.

The method uses a novel workflow capable of automatically generating tens of thousands of detailed fenestration solutions. Integrated simulations streamline the optimization process, ensuring full automation and eliminating the need for manual calculations. Unlike other ZEB optimization tools, this one is designed for practical use, enhancing productivity and accuracy in architecture and engineering offices by eliminating repetitive tasks and reducing decision-making biases.

The analysis of the optimization process demonstrates that the L-BFGS-B algorithm is not only a robust approach capable of consistently converging to solutions with similar properties across multiple executions but also significantly outperforms the Genetic Algorithm. It achieves this by improving both the quality of the solutions and reducing the variability of the best solutions obtained across all executions.

The **analysis of the best solutions** concludes that optimization offers valuable insights for ZEB design. Referring to the Iberian Peninsula, we can summarize:

- Shading system: Automated movable shading outperformed fixed shading in reducing heating and cooling demand and thermal discomfort across locations. The control system should adapt to the façade's solar orientation and climate variations.
- Glazing area: North and west glazing areas were minimized for natural light, while south glazing varied with shading type and composition, highlighting the tool's effectiveness in optimizing glazing size.
- Window composition: Regardless the shading type, frames with U-values above 1.19 should be excluded; argon gas, not air, should be used for glazing gaps, and low-emissivity treatments are recommended for north-facing façades, with coefficients between 0.05 and 0.07. When only fixed shading is used, glazing with solar control treatment is always required for west façades and almost always for south-facing façades, with SHGC values between 0.23 and 0.37.

- Cooling demand. In severe summer climates, all solutions exceeded allowable cooling demands, showing that even high-performing envelopes may struggle to meet the cooling limits of many international certification systems in warm climates. Data on the cooling demand of actual ZEBs in warm climates is scarce, making it difficult to benchmark these results and establish realistic thresholds. With climate change driving up cooling demands in buildings [95], [96], the need for rapid technological advancements to deliver more non-polluting energy is becoming urgent.

**Future research and limitations.**

This research focused on optimizing the transparent part of the building envelope, with ongoing efforts to extend optimization to the opaque surface and HVAC systems, for a broader range of ZEB-compliant solutions. Future work will integrate the tool into BIM systems, introduce new quality functions, such as daylighting and environmental impact, and optimize the ZEB envelope for climate change scenarios, ensuring the resilience of the proposed solutions. In all the tests conducted, a significant portion of the computation time was spent on Energy Plus simulations, which presents a notable limitation. To mitigate this, we propose integrating surrogate models into the optimization engine to dynamically learn simulation characteristics and minimize the need of Energy Plus executions. However, it is important to note that surrogate models increase overall complexity, as they directly influence the optimization process and require careful management and fine-tuning [97].

# CREDIT AUTHORSHIP CONTRIBUTION STATEMENT

**Rosana Caro**: Conceptualization, Methodology, Formal analysis, Investigation, Data curation, Writing-original draft, Writing-review & editing, Validation, Visualization, Supervision.

**Lorena Cruz**: Conceptualization, Methodology, Validation, Investigation, Resources, Data curation, Writing-original draft, Writing-review & editing, Supervision, Project administration, Funding acquisition.

**Pablo S Naharro**. Conceptualization, Methodology, Software, Validation, Formal analysis, Investigation, Data curation, Writing-original draft, Writing-review & editing.

**Santiago Muelas**. Conceptualization, Methodology, Validation, Formal analysis, Investigation, Data curation, Writing-original draft, Writing-review & editing, Supervision, Project administration

**Arturo Martínez**: Investigation, Formal analysis.

**Kevin King Sancho**. Software, Data curation.

**Elena Cuerda**. Conceptualization, Resources, Supervision, Project administration, Writing-review & editing, Funding acquisition.

**María del Mar Barbero-Barrera**. Conceptualization, Resources, Supervision, Project administration, Writing-review & editing, Funding acquisition.

**Antonio LaTorre**. Conceptualization, Funding acquisition, Methodology, Project administration, Supervision, Validation, Writing-review & editing.

# DECLARATION OF COMPETING INTEREST

The authors declare that they have no known competing financial interests or personal relationships that could have appeared to influence the work reported in this paper.

# FUNDING

This publication is part of the project PLEC2021007962, funded by the Spanish Ministry of Science and Innovation MCIN/AEI/10.13039/501100011033 and the European Union "NextGenerationEU"/PRTR.

# Appendix A. Rules for double glazing construction in dwellings of multi-family buildings

R1. All glazing in the same façade must have the same layered composition, as it will vary with its solar orientation in accordance with the following rules R7, R8 and R9. The graph below shows the criteria adopted to classify the solar orientation of facades (Fig. 18)

R2. Within the same building, all window frames must be made from the same material, have the same solar reflectance (colour), and meet the same air tightness class.

R3. The glazing can be filled with air or argon, and the gap width can be 6, 8, 10, 12, or 16 mm (Fig. 19)

R4. The outer glass (Glass 1) must be at least as thick as the inner glass (Glass 2). It may be one class[2] thicker than the inner glass.

R5. The glazing may or may not have a treatment for solar radiation control. If it does, the outer glass (Glass 1) must have a reduced Solar Transmittance ($T_{sol} < 0.54$).

R6. The glazing may or may not have a low-emissivity treatment. This treatment can be applied to either Glass 1 or Glass 2, but not to both.

R7. On north-facing façades ($\alpha < 60º$ or $\alpha \leqslant 300º$), glazing with solar control ($T_{sol} < 0.54$) is not permitted. The glazing may or may not have a low-emissivity treatment, which can be applied to either side #2 of Glass 1 or side #3 of Glass 2.

R8. On east and west-facing façades, the glazing may or may not have a reduced $T_{sol}$ and low-emissivity treatment. The latter can be applied to either side #2 of Glass 1 or side #3 of Glass 2.

R9. On south, southeast, and southwest-facing façades, the glazing may or may not have a reduced $T_{sol}$ and low-emissivity treatment. The latter can only be applied to side #2 of Glass 1.

---

[2] Classes are: 4, 6, 8 and 10 mm

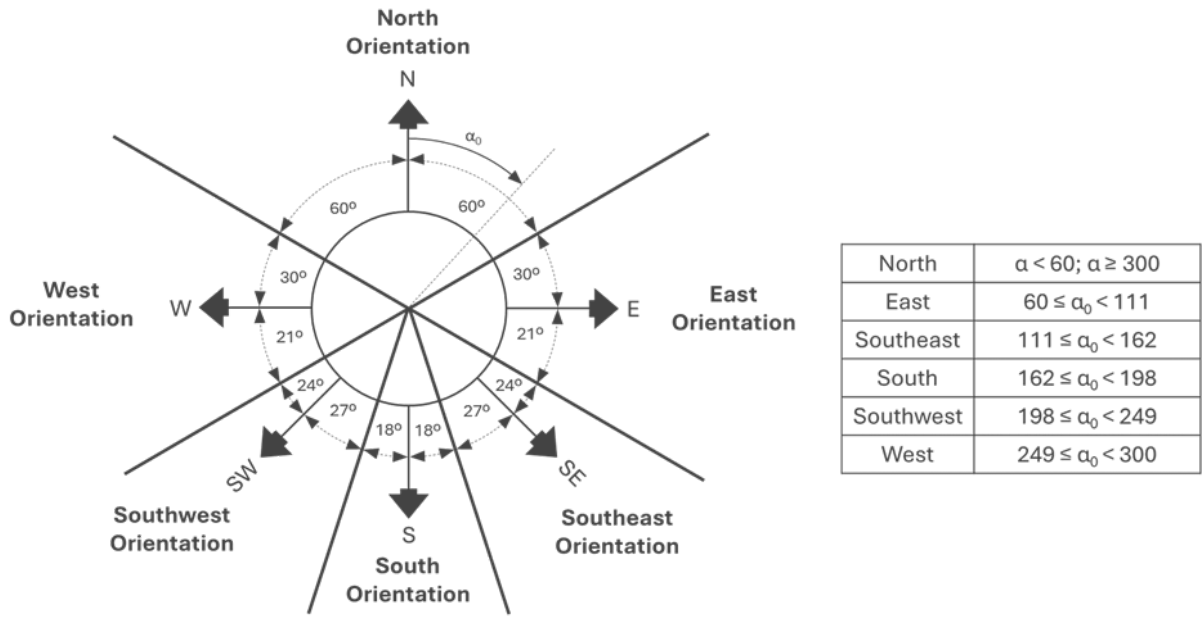

**Fig. 18**. Adopted criteria for solar orientation of façades

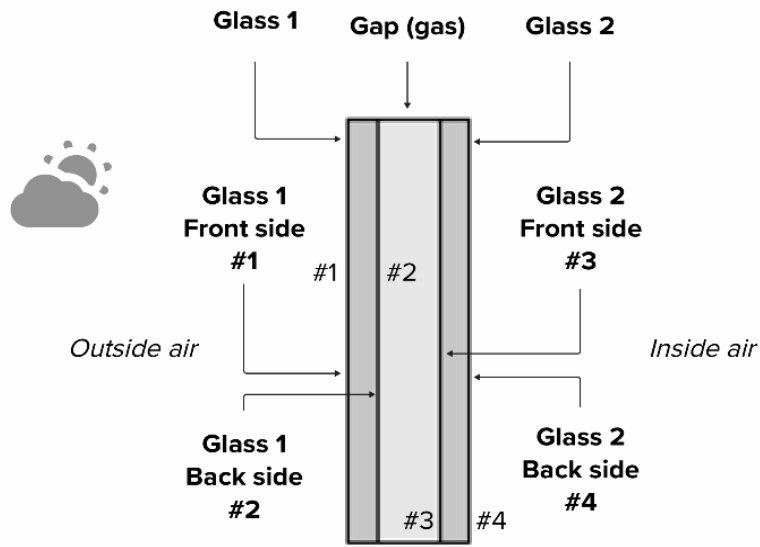

**Fig. 19**. Double glazing with the identification of their parts and the designation of their faces.

# Appendix B. Rules for sizing and positioning glazing in multi-family dwellings

R1. All windows are regular quadrilaterals.

R2. All glazing must fit within the designated area of the wall, defined by the red lines in Fig. 20

R3. The upper horizontal edge of any glazing must align with the upper boundary of the designated area (Fig. 20).

R4. The width of any glazing must be at least 0.60 meters[3] (Fig. 20).

R5. The horizontal axis of each glazing must not be more than 1.60 meters from the floor to ensure easy access to the handle[4] (Fig. 20).

R6. Windows must be centered on the wall, with their vertical axis aligned with the vertical axis of the designated area (Fig. 20).

R7. The total glazing surface in living rooms and bedrooms must be at least 12% of their usable floor area.

R8. Bathrooms and kitchens must each have exactly one window, with a fixed width of 0.60 meters.

R9. If a room has two exterior walls, glazing can be installed in only one of them due to furnishing constraints.

R10. In a bedroom or living room, glazing may cover up to the entire available area of the wall on which it is installed.

---

[3] According to rules R3, R4 and R5 the smallest window is 60 cm width and 100 cm high

[4] According to rules R3 and R5 the minimum height of any glazing is 1 m

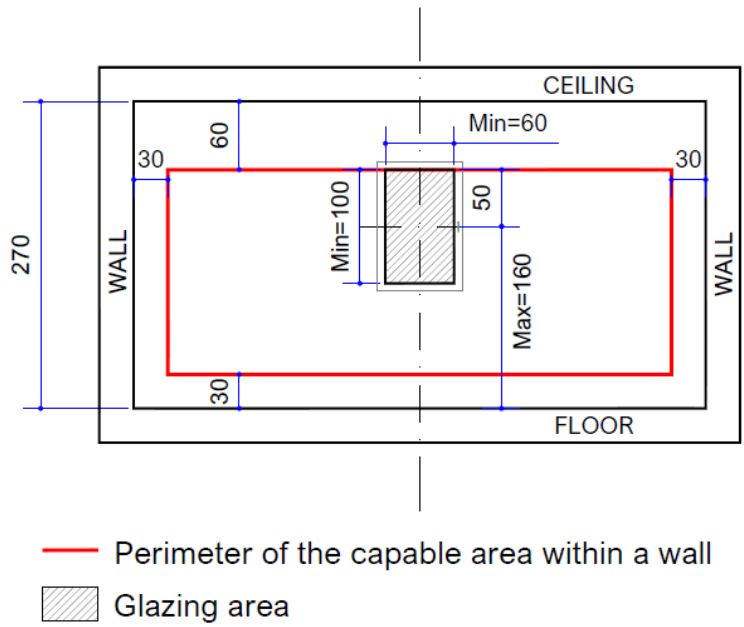

**Fig. 20**. Position of windows within the designated area of external walls (red perimeter). The figure also shows the smallest allowable window size based on the established rules. Dimensions are in centimeters (indicated by the blue lines).

## Appendix C. Rules for positioning and sizing of attached shading surfaces

R1. Overhangs and fins can be installed on windows on the North, South, and West facades.

R2. Overhangs must extend at a 90º angle relative to the window (Fig.21)

R3. Fins must be positioned at a 90º angle to the window in the plan view.

R4. Overhangs are only allowed on the top window frame, not the bottom

R5. Bottom fin extensions are not allowed.

R6. Overhangs and fins must be aligned with the outer edge of the window frame, with no gap.

R7. Depth and extension of shading devices are measured in 0.1 m increments.

R8. While the dimensions for west-facing windows are fixed, the depths and extensions of their shading devices can vary.

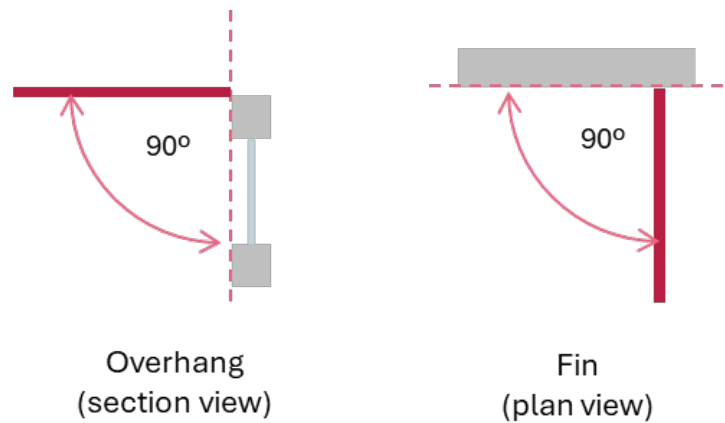

**Fig. 21**. Overhangs tilt angle (a) and fins rotation angle (b)